\documentclass[runningheads]{llncs}
\usepackage{graphicx}
\usepackage{comment}
\usepackage{amsmath,amssymb} 
\usepackage{color}
\usepackage{times}
\usepackage{epsfig}
\usepackage[font=scriptsize]{subfig}
\usepackage{enumitem}
\usepackage{tabularx}
\usepackage[flushleft]{threeparttable} 
\usepackage[export]{adjustbox}
\usepackage{booktabs}
\usepackage[dvipsnames, table]{xcolor}
\usepackage[toc,page]{appendix}

\usepackage{hyperref}

\hypersetup{
  colorlinks, linkcolor=red
}
\makeatletter
\g@addto@macro{\UrlBreaks}{\UrlOrds}
\makeatother

\newcommand{\IR}{\mathbb{R}}

\newcommand{\EX}{\mathcal{X}}
\newcommand{\EY}{\mathcal{Y}}
\newcommand{\EZ}{\mathcal{Z}}
\newcommand{\EP}{\mathcal{P}}
\newcommand{\EL}{\mathcal{L}}
\newcommand{\EE}{\mathcal{E}}

\newcommand{\etal}{\textit{et al}.}
\newcommand{\ie}{\textit{i}.\textit{e}., }

\definecolor{myGreen}{rgb}{0.17,0.64,0.37}
\definecolor{myRed}{rgb}{0.95,0.1,0.10}

\newlength{\tempdima}
\newcommand{\rowname}[1]
{\rotatebox{90}{\makebox[\tempdima][c]{\textbf{#1}}}}

\makeatletter
\renewcommand\paragraph{\@startsection{paragraph}{4}{\z@}%
                                      {\parskip}
                                      {-1em}%
                                      {\normalfont\normalsize\bfseries}}
\makeatother

\begin{document}
\pagestyle{headings}
\mainmatter
\def\ECCVSubNumber{6985}  

\title{Child Face Age-Progression via Deep Feature Aging} 

\author{Debayan Deb\and
Divyansh Aggarwal\and
Anil K. Jain}
\institute{Michigan State University, East Lansing, MI, USA\\
\email{\{debdebay,aggarw49,jain\}@cse.msu.edu}}
\maketitle

\begin{abstract}
Given a gallery of face images of missing children, state-of-the-art face recognition systems fall short in identifying a child (probe) recovered at a later age. We propose a feature aging module that can age-progress deep face features output by a face matcher. In addition, the feature aging module guides age-progression in the image space such that synthesized aged faces can be utilized to enhance longitudinal face recognition performance of any face matcher without requiring any explicit training. For time lapses larger than 10 years (the missing child is found after 10 or more years), the proposed age-progression module improves the closed-set identification accuracy of FaceNet from 16.53\% to 21.44\% and CosFace from 60.72\% to 66.12\% on a child celebrity dataset, namely ITWCC. The proposed method also outperforms state-of-the-art approaches~\cite{look_through_elapse,decorrelated} with a rank-1 identification rate of 95.91\%, compared to 94.91\%, on a public aging dataset, FG-NET, and 99.58\%, compared to 99.50\%, on CACD-VS. These results suggest that aging face features enhances the ability to identify young children who are possible victims of child trafficking or abduction.
\end{abstract}

\section{Introduction}
Human trafficking is one of the most adverse social issues currently faced by countries worldwide. According to the United Nations Children's Fund (UNICEF) and the Inter-Agency Coordination Group against Trafficking (ICAT), 28\% of the identified victims of human trafficking globally are children\footnote{\scriptsize The United Nations Convention on the Rights of the Child defines a child as “a human being below the age of
18 years unless under the law applicable to the child, majority is attained earlier~\cite{Child}.}~\cite{UNICEF}. 
The actual number of missing children is much more than these official statistics as only a limited number of cases are reported because of the fear of traffickers, lack of information, and mistrust of authorities.

Face recognition is perhaps the most promising biometric technology for recovering missing children, since parents and relatives are more likely to have a lost child's face photograph than other biometric modalities such as fingerprint or iris\footnote{\scriptsize Indeed, face is certainly not the only biometric modality for identification
of lost children. Sharbat Gula, first photographed in 1984 (age 12) in a refugee
camp in Pakistan, was later recovered via iris recognition at the age of 30 from a remote part of Afghanistan in 2002~\cite{sherbat}.}.
While Automated Face Recognition (AFR) systems have been able to achieve high identification rates in several domains~\cite{cosface,facenet,nist}, their ability to recognize children as they age is still limited.

\begin{figure}[!t]
    \centering
    \scriptsize
    \footnotesize{26 years\hspace{3em}6 years\hspace{3em}26 years\hspace{4em}16 years\hspace{3em}5 years\hspace{3em}16 years}\\
        \includegraphics[width=0.15\linewidth]{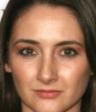}
        \includegraphics[width=0.15\linewidth]{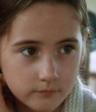}
        \includegraphics[width=0.15\linewidth]{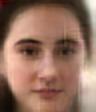}
    \hspace{2em}
        \includegraphics[width=0.15\linewidth]{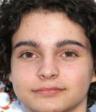}
        \includegraphics[width=0.15\linewidth]{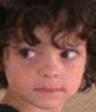}
        \includegraphics[width=0.15\linewidth]{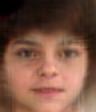}\\
        \footnotesize{\hspace{6em}\textcolor{Red}{\textbf{0.33}}\hspace{4em}\textcolor{Green}{\textbf{0.39}}\hspace{12em}\textcolor{Red}{\textbf{0.34}}\hspace{4em}\textcolor{Green}{\textbf{0.42}\hspace{7em}}}\\[-1em]
    \hspace{2em} (a) Hannah Taylor Gordon \hspace{10em} (b)  Max Burkholder \hspace{2em}
\caption{Cols 2 $\&$ 5: Face images of two child celebrities enrolled in the gallery. Cols 1 $\&$ 4: The celebrities’ probe images at different ages (denoted above each photo). As the children grow older, a state-of-the-art face
matcher, CosFace~\cite{cosface}, fails to match the enrolled image of the same child (highlighted in red). Cols 3 $\&$ 6: With the proposed aging scheme, we synthesize an age-progressed face image at the probe's age which matches the probe image with higher similarity score (cosine similarity scores given below each score range from $[-1, 1]$ and threshold for CosFace is $0.35$ at $0.1\%$ FAR).}
\label{fig:frontpage}
\end{figure}

A human face undergoes various temporal changes, including skin texture, weight, facial hair, etc. (see Figure~\ref{fig:frontpage})~\cite{anatomy_face,facialstructure}. Several studies have analyzed the extent to which facial aging affects the performance of AFR. Two major conclusions can be drawn based on these studies: (i) Performance decreases with an increase in time lapse between subsequent image acquisitions~\cite{klare,deb_adult,nist_2018}, and (ii) performance degrades more rapidly in the case of younger individuals than older individuals~\cite{nist_2018,deb_child}.
Figure~\ref{fig:frontpage} illustrates that a state-of-the-art face matcher (CosFace) fails when it comes to matching an enrolled child in the gallery with the corresponding probe over large time lapses. Thus, it is essential to enhance the longitudinal performance of AFR systems, especially when the child is enrolled at a young age. 


\footnotetext{\scriptsize The award-winning 2016 movie, Lion, is based on the true story of Saroo Brierley~\cite{saroo_movie}.} 

\subsection{Limitations of Existing Solutions}
Prior studies on face recognition under aging, both for adults and children, explored both \emph{generative} and \emph{discriminative} models. Discriminative approaches focus on \emph{age-invariant} face recognition under the assumption that age and identity related components can be separated~\cite{lfcnn,discriminative,aecnn,oecnn,look_through_elapse,decorrelated}. By separating age-related components, only the identity-related information is used for face matching. Since age and identity can be correlated in the feature space, the task of disentangling them from face embeddings is not only difficult but can also be detrimental to AFR performance~\cite{hill,attributes}. 

Given a probe face image, generative models can synthesize face images that can either predict how the person will look over time (age progression) or estimate how he looked in previous years (age regression) by utilizing Generative Adversarial Networks (GANs)~\cite{geng,caae,ipcgan,cgan,hongyu,lanitis}. Prior studies are primarily motivated to enhance the visual quality of the age progressed or regressed face images, rather than enhancing the face recognition performance.

A majority of the prior studies on \emph{cross-age face recognition}\footnote{\scriptsize Face matching or retrieval under aging changes}~\cite{lfcnn,aecnn,oecnn,hongyu,decorrelated,look_through_elapse} evaluate the performance of their models on longitudinal face datasets, such as MORPH ($13,000$ subjects in the age range of $16$-$77$ years) and CACD ($2,000$ subjects in the age range of $16$-$62$ years), which do not contain face images of young subjects. Indeed, some benchmark face datasets such as FG-NET ($82$ subjects in the age range of $0$-$69$ years) do include a small number of children, however, the associated protocol is based on matching~\emph{all possible comparisons} for all ages, which does not explicitly provide child-to-adult matching performance. Moreover, earlier longitudinal studies employ \emph{cross-sectional} techniques, where the temporal performance is analyzed according to differences between age groups~\cite{decorrelated,caae,bereta}. In cross-sectional or cohort-based approaches, which age groups or time lapses are evaluated is often arbitrary and
varies from one study to another, thereby, making comparisons between studies difficult~\cite{yoon,lacey_adult}. For these reasons, since facial aging is longitudinal by nature, cross-sectional analysis is not the correct model for exploring aggregated effects~\cite{yoon,lacey,nist_irex_report}.
 


\subsection{Our Contributions}
We propose a feature aging module that learns a projection in the deep feature space. In addition, the proposed feature aging module can guide the aging process in the image space such that we can synthesize visually convincing face images for any specified target age (not age-groups). These aged images can be directly used by any commodity matcher for enhanced longitudinal performance. Our empirical results show that the proposed feature aging module, along with the age-progressed\footnote{\scriptsize Though our module is not strictly restricted to age-progression, we use the word progression largely because in the missing children scenario the gallery would generally be younger than the probe. Our module does both age-progression and age-regression when we benchmark our performance on public datasets.} images, can improve the longitudinal face recognition performance of three face matchers (FaceNet~\cite{facenet}, CosFace~\cite{cosface}, and a commercial-off-the-shelf (COTS) matcher)\footnote{\scriptsize Since we do not have access to COTS features, we only utilize COTS as a baseline.} for matching children as they age.

The specific contributions of the paper are as follows:
\begin{itemize}
\item A feature aging module that learns to traverse the deep feature space such that the identity of the subject is preserved while only the age component is progressed or regressed in the face embedding for enhanced longitudinal face recognition performance.

    \item An image aging scheme guided by our feature aging module to synthesize an age-progressed or age-regressed face image at any specified target age. The aged face images can enhance face recognition performance for matchers unseen during training.
    \item With the proposed age-progression module, rank-1 identification rates of a state-of-the-art matcher, CosFace~\cite{cosface}, increase from $91.12\%$ to $94.24\%$ on CFA (a child face aging dataset), and $38.16\%$ to $55.30\%$ on ITWCC~\cite{ITWCC-D1} (a child celebrity dataset). In addition, the proposed module boosts accuracies from $94.91\%$ and $99.50\%$ to $95.91\%$ and $99.58\%$ on FG-NET and CACD-VS respectively, which are the two public face aging benchmark datasets~\cite{fgnet,cacd}\footnote{\scriptsize We follow the exact protocols provided with these datasets. We open-source our code for reproducibility: [\textit{url omitted for blind review}].}.
\end{itemize}

\section{Related Work}
\begin{table*}[!t]
\scriptsize
\captionsetup{font=scriptsize}
\setlength{\tabcolsep}{3pt}
\caption{Face aging datasets. Datasets below solid line includes longitudinal face images of children.}
\centering
\begin{threeparttable}
\begin{tabularx}{\linewidth}{l l l l l l l l}	
\noalign{\hrule height 1.0pt}
Dataset & No. of Subjects & No. of Images & No. Images / Subject & Age Range (years) & Avg. Age (years)  & Public\protect\footnotemark\\
  \noalign{\hrule height 0.3pt}
  MORPH-II~\cite{morph} &13,000 & 55,134 & 2-53 (avg. 4.2) &16-77 & 42 & Yes\\
  \hline
    CACD~\cite{cacd} &  2,000 & 163,446   & 22-139 (avg. 81.7) & 16-62 & 31 & Yes\\\hline
    FG-NET~\cite{fgnet} & 82 & 1,002 & 6-18 (avg. 12.2) & 0-69 & 16 &  Yes\\
    \hline
    UTKFace~\cite{caae}\tnote{$\dagger$} & N/A & 23,708 & N/A & 0-116 & 33 & Yes\\
    \hline
    ITWCC~\cite{ITWCC} & 745 & 7,990 & 3-37 (avg. 10.7) & 0-32 & 13 & No\tnote{$\dagger\dagger$}\\
    \hline 
    CLF~\cite{deb_child} & 919 & 3,682 & 2-6 (avg. 4.0) & 2-18 & 8 & No\tnote{$\dagger\dagger$}\\ \hline
    CFA & 9,196 & 25,180 & 2-6 (avg. 2.7) & 2-20 & 10 & No\tnote{$\dagger\dagger$}\\
    \noalign{\hrule height 0.8pt}

\end{tabularx}
\begin{tablenotes}\scriptsize
\item[$\dagger$]\hspace{0.2em} Dataset does not include subject labels; Only a collection of face images along with the corresponding ages.
\item[$\dagger\dagger$] Concerns about privacy issues are making it extremely difficult for researchers to place the child face images in public domain. 
\end{tablenotes}
\end{threeparttable}
\label{tab:allDatasets}
\end{table*}

\footnotetext{\scriptsize MORPH: \url{https://bit.ly/31P6QMw}, CACD: \url{https://bit.ly/343CdVd}, FG-NET: \url{https://bit.ly/2MQPL0O}, UTKFace:~\url{https://bit.ly/2JpvX2b}}

\begin{table*}[!t]
\scriptsize
\setlength{\tabcolsep}{5pt}
\captionsetup{font=scriptsize}
\caption{Related work on cross-age face recognition. Studies below bold line deal with children.}
\centering
\begin{threeparttable}
\renewcommand{\arraystretch}{1.5}
\begin{tabularx}{\textwidth}{l l X X}
\noalign{\hrule height 1.0pt}
\textbf{Study} & Objective & Dataset & Age groups or range (years)\\
  \noalign{\hrule height 1.0pt}
  \textbf{Yang~\etal~\cite{hongyu}}\small{\tnote{*}} & Age progression of face images & MORPH-II, CACD & 31-40, 41-50, 50+\\
    \hline
  \textbf{Wang~\etal~\cite{decorrelated}}\small{\tnote{*}} & Decomposing age and identity & MORPH-II, FG-NET, CACD & 0-12, 13-18, 19-25, 26-35, 36-45, 46-55, 56-65, 65+\\
    \hline
 \textbf{Best-Rowden~\etal~\cite{lacey_adult}} & Model for change in genuine scores over time & PCSO, MSP & 18-83\\ \hline
  \textbf{Ricanek~\etal~\cite{ITWCC}} &  Face comparison of infants to adults & ITWCC & 0-33\\
    \hline
\textbf{Deb~\etal~\cite{deb_child}} & Feasibility study of AFR for children & CLF & 2-18\\
\hline
\textbf{This study} & Aging face features for enhanced AFR for children & CFA, ITWCC, FG-NET, CACD & 0-18\\
\noalign{\hrule height 1.0pt}
\end{tabularx}
\begin{tablenotes}\scriptsize
\item[*] Study uses cross-sectional model (ages are partitioned into age groups) and not the correct longitudinal model~\cite{deb_adult},~\cite{yoon}.
\end{tablenotes}
 \end{threeparttable}
\label{tab:related}
\end{table*}

\subsection{Discriminative Approaches}
Approaches prior to deep learning leveraged robust local descriptors~\cite{hfa,mefa,hierarchical,discriminative,ling} to tackle recognition performance degradation due to face aging. Recent approaches focus on age-invariant face recognition by attempting to discard age-related information from deep face features~\cite{lfcnn,aecnn,coupled,look_through_elapse,decorrelated}. All these methods operate under two critical assumptions: (1) age and identity related features can be disentangled, and (2) the identity-specific features are adequate for face recognition performance. Several studies, on the other hand, show that age is indeed a major contributor to face recognition performance~\cite{hill,attributes}. Therefore, instead of completely discarding age factors, we exploit the age-related information to progress or regress the deep feature directly to the desired age.

\subsection{Generative Approaches}
Ongoing studies leverage Conditional Auto-encoders and Generative Adversarial Networks (GANs) to synthesize faces by automatically learning aging patterns from face aging datasets~\cite{recurrent,caae,look_through_elapse,hongyu,cgan,ipcgan,unsang}. The primary objective of these methods is to synthesize visually realistic face images that appear to be age progressed, and therefore, a majority of these studies do not report the recognition performance of the synthesized faces.


\subsection{Face Aging for Children}
Best-Rowden~\etal \ studied face recognition performance of
newborns, infants, and toddlers (ages 0 to 4 years) on 314 subjects acquired over a time lapse of only one
year~\cite{lacey}. Their results showed a True Accept Rate (TAR) of 47.93\% at 0.1\% False Accept Rate (FAR) for an age group of [0, 4] years for a commodity face matcher. Deb~\etal\ fine-tuned FaceNet~\cite{facenet} to achieve a rank-1 identification accuracy of $77.86\%$ for a time lapse between the gallery and probe image of 1 year. Srinivas~\etal\ showed that the rank-1 performance of state of the art commercial face matchers on longitudinal face images from the In-the-Wild Child Celebrity (ITWCC)~\cite{ITWCC-D1} dataset ranges from $42\%$ to $78\%$ under the youngest to older protocol. These studies primarily focused on evaluating the longitudinal face recognition performance of state-of-the-art face matchers rather than proposing a solution to improve face recognition performance on children as they age.
Table~\ref{tab:allDatasets} summarizes longitudinal face datasets that include children and Table~\ref{tab:related} shows related work in this area.

\section{Proposed Approach}
Suppose we are given a pair of face images $(x, y)$
for a child acquired at ages $e$ (enrollment) and $t$ (target), respectively.
Our goal is to enhance the ability of state-of-the-art face recognition systems to match $x$ and $y$ when ($e\ll t$).

We propose a feature aging module that learns a projection of the deep features in a lower-dimensional space which can directly improve the accuracy of face recognition systems in identifying children over large time lapses. The feature aging module guides a generator to output aged face images that can be used by any commodity face matcher for enhancing cross-age face recognition performance.

\subsection{Feature Aging Module (FAM)}
It has been found that age-related components are highly coupled with identity-salient features in the latent space~\cite{hill,attributes}. That is, the age at which a face image is acquired can itself be an intrinsic feature in the latent space. Instead of disentangling age-related components from identity-salient features, we would like to automatically learn a projection within the face latent space.

Assume $x\in\EX$ and $y\in\EY$ where $\EX$ and $\EY$ are two face domains when images are acquired at ages $e$ and $t$, respectively. Face manipulations shift images in the domain of $\EX$ to $\EY$ via,
\begin{align}
    \label{eq_1}
    \hat{y} = \mathcal{F}(x, e, t)
\end{align}
where, $\hat{y}$ is the output image and $\mathcal{F}$ is the operator that changes $x$ from $\EX$ to $\EY$.
Domains $\EX$ and $\EY$ generally differ in factors other than aging, such as noise, quality and pose. Therefore, $\mathcal{F}$ can be highly complex. We can simplify $\mathcal{F}$ by modeling the transformation in the deep feature space by defining an operator $\mathcal{F'}$ and rewrite equation~\ref{eq_1}
\begin{align}
    \hat{y} = \mathcal{F'}(\psi(x), e, t)
\end{align}
where $\psi(x)$ is an encoded feature in the latent space. Here, $\mathcal{F'}$ learns a projection in the feature space that shifts an image $x$ in $\EX$ to $\EY$ and therefore, `ages' a face feature from $e$ to $t$. Since face representations lie in $d$-dimensional Euclidean space\footnote{\scriptsize Assume these feature vectors are constrained to lie in a $d$-dimensional hypersphere,~\ie, $||\psi(x)||_2^2 = 1$.}, $\EZ$ is highly linear. Therefore, $\mathcal{F'}$ is a linear shift in the deep space, that is,
\begin{align}
    \label{eq_3}
    \hat{y} = \mathcal{F'}(\psi(x), e, t) = W\times(\psi(x)\oplus e\oplus t)+b
\end{align}
where $W\in\IR^{d\times d}$ and $b\in\IR^{d}$ are learned parameters of $\mathcal{F'}$ and $\oplus$ is concatenation. The scale parameter, $W$ allows each feature to scale differently given the enrollment and target ages since not all face features change drastically during the aging process (such as eye color).

\subsection{Image Generator}
While, FAM can directly output face embeddings that can be used for face recognition, FAM trained on the latent space of one matcher may not generalize to feature spaces of other matchers. Still, the feature aging module trained for one matcher can guide the aging process in the image space. In this manner, an aged image can be directly used to enhance cross-age face recognition performance for any commodity face matcher without requiring any re-training. The aged images should (1) enhance identification performance of white-box and black-box\footnote{\scriptsize White-box matcher is a face recognition system utilized during the training of FAM module, whereas, \emph{black-box} matcher is a commodity matcher that is typically not used during training.} face matchers under aging, and (2) appear visually realistic to humans. 

\section{Learning Face Aging}
\begin{figure*}[!t]
    \centering
    \captionsetup{font=scriptsize}
    \includegraphics[width=0.7\linewidth]{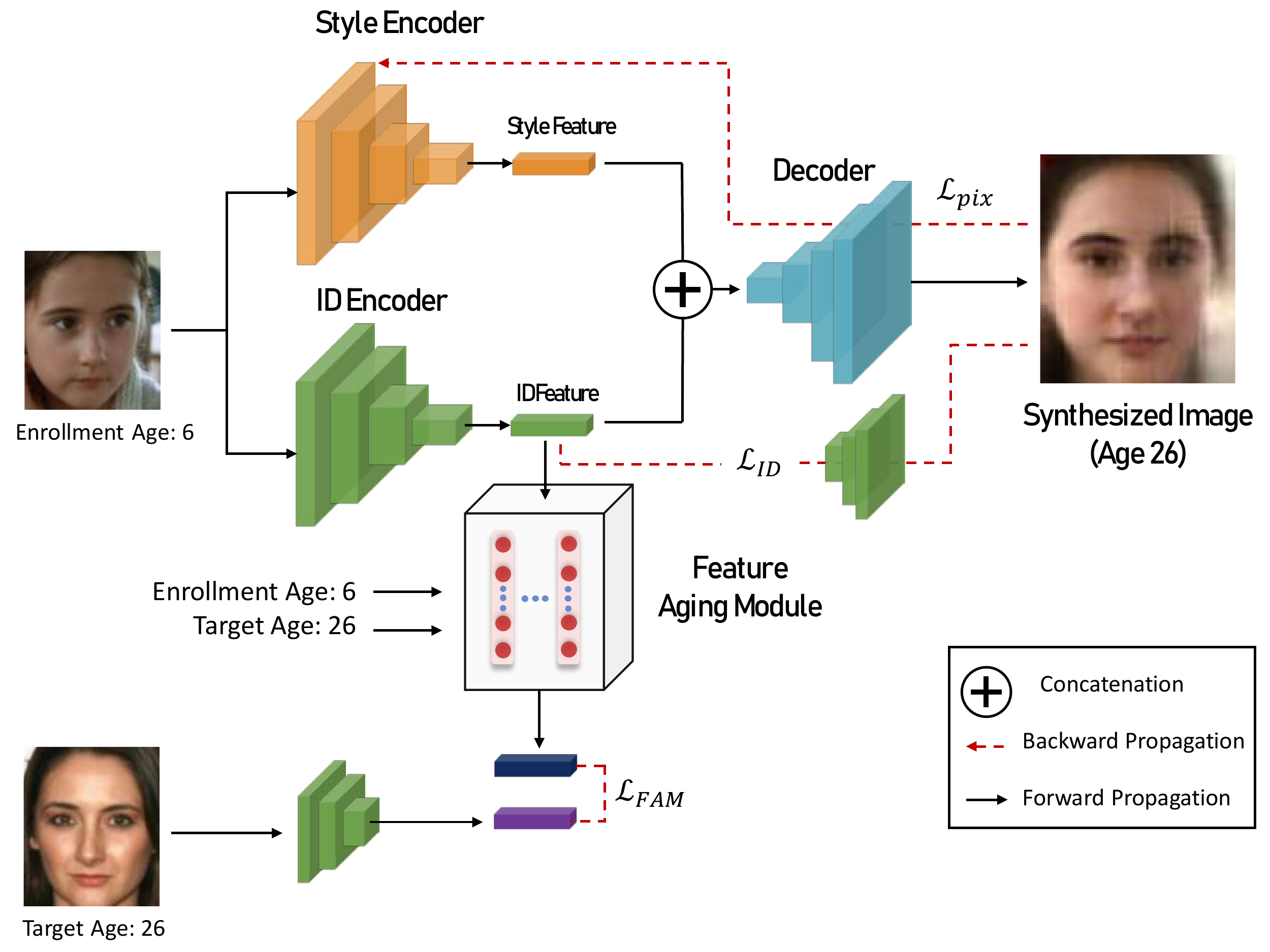}
    \caption{Overview of the proposed child face aging training framework. The feature aging module guides the decoder to synthesize a face image to any specified target age while the style-encoder injects style from the input probe into the synthesized face. For simplicity, we omit the total variation loss ($\EL_{TV}$).}
    \label{fig:overview}
\end{figure*}

Our proposed framework for face aging comprises of three modules, (i) feature aging module (FAM), (ii) style-encoder, and (iii) decoder. For training these three modules, we utilize a fixed ID encoder ($\EE_{ID}$) that outputs a deep face feature from an input face image. An overview of our proposed aging framework is given in Figure~\ref{fig:overview}.

\subsection{Feature Aging Module (FAM)} This module consists of a series of fully connected layers that learn the scale $W$ and bias $b$ parameters in Equation~\ref{eq_3}. For training, we pass a genuine pair of face features $(\psi(x),\psi(y))$ extracted from an identity encoder ($\EE_{ID}$), where $x$ and $y$ are two images of the same person acquired at ages $e$ and $t$, respectively. In order to ensure that the identity-salient features are preserved and the synthesized features are age-progressed to the desired age $t$, we train FAM via a mean squared error (MSE) loss which measures the quality of the predicted features:
\begin{align}
    \EL_{FAM} = \frac{1}{|\EP|} \sum_{(i, j)\in \EP} || \mathcal{F'}(\psi(x), e, t) - \psi(y)||_2^2,
\end{align}
where $\EP$ is the set of all genuine pairs. After the model is trained, FAM can progress a face feature to the desired age. Our experiments show that $L_{FAM}$ forces the FAM module to retain all other covariates (such as style and pose) in the predicted features from the input features.

\subsection{Style-Encoder} The ID encoder ($\EE_{ID}$) encodes specific information pertaining to the identity of a person's face. However, a face image comprises of other pixel-level residual features that may not relate to a person's identity but are required for enhancing the visual quality of the synthesized face image which we refer to as $style$. Directly decoding a face feature vector into an image can hence, severely affect the visual quality of the synthesized face. Therefore, we utilize a style-encoder ($\EE_{style}$) that takes a face image, $x$, as an input and outputs a $k$-dimensional feature vector that encodes style information from the image.

\subsection{Decoder} In order to synthesize a face image, we propose a decoder $\mathcal{D}$ that takes as input a style and an ID vector obtained from $\EE_{style}$ and $\EE_{ID}$, respectively, and outputs an image $y$.

\paragraph{Identity-Preservation} Our goal is to synthesize aged face images that can benefit face recognition systems. Therefore, we need to constrain the decoder $\mathcal{D}$ to output face images that can be matched to the input face image. To this end, we adopt an identity-preservation loss to minimize the distance between the input and synthesized face images of the decoder in the $\EE_{ID}$'s latent space,
\begin{align}
    \EL_{ID} = \sum_{i=0}^{n}||\EE_{ID}(\mathcal{D}(\EE_{style}(x_i), \EE_{ID}(x_i))) - \EE_{ID}(x_i)||^2_2
\end{align}
where $x_i$ are samples in the training set.

\paragraph{Pixel-level Supervision} In addition to identity-preservation, a pixel-level loss is also adopted to maintain the consistency of low-level image content between the input and output of the decoder,
\begin{align}
    \EL_{pix} = \sum_{i=0}^{n}||\mathcal{D}(\EE_{style}(x_i), \EE_{ID}(x_i)) - x_i||_1
\end{align}
In order to obtain a smooth synthesized image devoid of sudden changes in high-frequency pixel intensities, we regularize the total variation in the synthesized image,
\begin{align}
    \EL_{TV} = \sum_{i=0}^{n}\left[\sum_{r,c}^{H, W}\left[\left(x_{i_{r+1, c}} - x_{i_{r,c}}\right)^2 + \left(x_{i_{r, c+1}} - x_{i_{r,c}}\right)^2\right]\right]
\end{align}

For training $\mathcal{D}$, we do not require the aged features predicted by the feature aging module, since we enforce the predicted aged features to reside in the latent space of $\EE_{ID}$. During training we input a face feature extracted via $\EE_{ID}$ directly to the decoder. In the testing phase, we can use either a $\EE_{ID}$ face feature or an aged face feature from FAM to obtain a face image. 
Our final training objective for the style-encoder and decoder is,
\begin{align}
\EL_{(\EE_{style}, \mathcal{D})} = \lambda_{ID}\EL_{ID} + \lambda_{pix}\EL_{pix} + \lambda_{TV}\EL_{TV}
\end{align}
where $\lambda_{ID},~\lambda_{pix},~\lambda_{TV}$ are the hyper-parameters that control the relative importance of every term in the loss.

We train the Feature Aging Module and the Image Generator in an end-to-end manner by minimizing $\EL_{(\EE_{style}, \mathcal{D})}$ and $\EL_{FAM}$.

\section{Implementation Details}
\paragraph{Feature Aging Module} For all the experiments, we stack two fully connected layers and set the output of each layer to be of the same $d$ dimensionality as the ID encoder's feature vector.


We train the proposed framework for 200,000 iterations with a batch size of 64 and a learning rate of $0.0002$ using Adam optimizer with parameters $\beta_1 = 0.5, \beta_2 = 0.99$. In all our experiments, $k=32$. Implementations are provided in the supplementary materials.

\paragraph{ID Encoder} For our experiments, we employ 3 pre-trained face matchers\footnote{\scriptsize Both the open-source matchers and the COTS matcher achieve 99\% accuracy on LFW under LFW protocol.}. Two of them, FaceNet~\cite{facenet} and CosFace~\cite{cosface}, are publicly available. FaceNet is trained on VGGFace2 dataset~\cite{vggface2} using the Softmax+Center Loss~\cite{facenet}. CosFace is a 64-layer residual network~\cite{sphereface} and is trained on MS-ArcFace dataset~\cite{arcface} using AM-Softmax loss function~\cite{arcface}. Both matchers extract a $512$-dimensional feature vector. We also evaluate results on a commercial-off-the-shelf face matcher, COTS\footnote{\scriptsize This particular COTS utilizes CNNs for face recognition and has been used for identifying children in prior studies~\cite{ITWCC-D1,deb_child}. COTS is one of the top performers in the NIST Ongoing Face Recognition Vendor Test (FRVT)~\cite{nist_2018}.}. This is a closed system so we do not have access to its feature vector and therefore only utilize COTS as a baseline.

\section{Experiments}

\subsection{Cross-Age Face Recognition}
\label{sec:cafr}

\begin{figure}[!t]
\captionsetup{font=scriptsize}
\scriptsize
\centering
\begin{tabular}{@{}c@{ }c@{ }c@{ }c@{ }c@{ }}
5 years & 6 years & 8 years & 11 years\\

\includegraphics[width=0.11\linewidth]{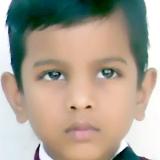}&
\includegraphics[width=0.11\linewidth]{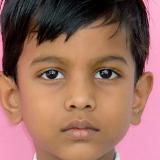}&
\includegraphics[width=0.11\linewidth]{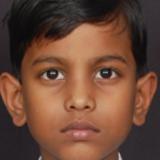}&
\includegraphics[width=0.11\linewidth]{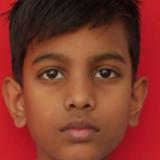}
\end{tabular}\hspace{2em}
\begin{tabular}{@{}c@{ }c@{ }c@{ }c@{ }c@{ }}
3 years & 5 years & 12 years & 13 years\\
\includegraphics[width=0.11\linewidth]{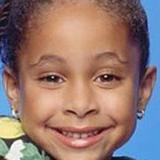}&
\includegraphics[width=0.11\linewidth]{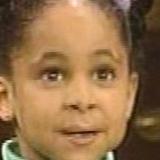}&
\includegraphics[width=0.11\linewidth]{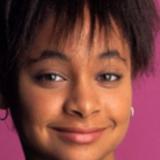}&
\includegraphics[width=0.11\linewidth]{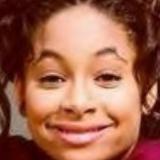}\end{tabular}\\
\textbf{\scriptsize (a) CFA}\hspace{20em}
\textbf{\scriptsize (b) ITWCC~\cite{ITWCC-D1}}
\caption{Examples of longitudinal face images from (a) CFA and (b) ITWCC~\cite{ITWCC-D1} datasets. Each row consists of images of one subject; age at image acquisition is given above each image}%
\label{fig:Dataset_examples}
\end{figure}

\begin{table*}[t]
\centering
\scriptsize
\captionsetup{font=scriptsize}
\setlength{\tabcolsep}{4pt}
\caption{Rank-1 identification accuracy on two child face datasets, CFA and ITWCC~\cite{ITWCC-D1}, when the time gap between a probe and its true mate in the gallery is larger than 5 years and 10 years, respectively. The proposed aging scheme (in both the feature space as well as the image space) improves the performance of FaceNet and CosFace on cross-age face matching. We also report the number of probes (P) and gallery sizes (G) for each experiment.}
\begin{threeparttable}
\begin{tabular}{l||c|c||c|c}
\noalign{\hrule height 1.0pt}
                & \multicolumn{2}{c||}{\textbf{CFA (Constrained)}}
                & \multicolumn{2}{c}{\textbf{ITWCC (Semi-Constrained)~\cite{ITWCC-D1}}}\\ \hline
\textbf{Method}  & \textbf{Closed-set} & \textbf{Open-set\tnote{$\dagger$}} & \textbf{Closed-set} & \textbf{Open-set\tnote{$\dagger$}}  \\ 
\noalign{\hrule height 0.3pt}
                           & \textbf{Rank-1 }             & \textbf{Rank-1 @ 1\% FAR}
                           &  \textbf{Rank-1}              & \textbf{Rank-1 @ 1\% FAR}    \\
\noalign{\hrule height 0.1pt}
                           & {P: 642, G: 2213}             & P: 3290, G: 2213
                           & P: 611, G: 2234             & P: 2849, G: 2234  \\ 
\noalign{\hrule height 0.8pt}
COTS~\cite{ITWCC-D1}  &  91.74 & 91.58 & 53.35 & 16.20 \\\noalign{\hrule height 0.8pt}
FaceNet~\cite{facenet}  & 38.16 & 36.76 & 16.53 & 16.04 \\ (w/o Feature Aging Module) & & & &\\ \hline
FaceNet & \textbf{55.30} & \textbf{53.58} &  \textbf{21.44} & \textbf{19.96}\\ (with Feature Aging Module)& & & &\\
\noalign{\hrule height 0.8pt}
CosFace~\cite{cosface}  & 91.12 & 90.81 &  60.72   &  22.91 \\ (w/o Feature Aging Module)  & & & &\\ \hline
CosFace  & \textbf{94.24} & \textbf{94.24} & \textbf{66.12} &  \textbf{25.04}\\ (with Feature Aging Module) & & & &\\
\noalign{\hrule height 0.8pt}
CosFace (Image Aging) & \textbf{93.18} & \textbf{92.47} &  \textbf{64.87} &  \textbf{23.40}\\
\noalign{\hrule height 0.8pt}
\end{tabular}
\begin{tablenotes}\scriptsize
\item[$\dagger$]A probe is first claimed to be present in the gallery. We accept or reject this claim based on a  pre-determined threshold @ $1.0\%$ FAR (verification). If the probe is accepted, the ranked list of
gallery images which match the probe with similarity scores
above the threshold are returned as the candidate list (identification).
\end{tablenotes}
\end{threeparttable}
\label{tab:results}
\end{table*}
\begin{figure*}
\centering
\scriptsize
\captionsetup{font=scriptsize}
    \subfloat[\scriptsize CFA]{\includegraphics[width=0.48\linewidth]{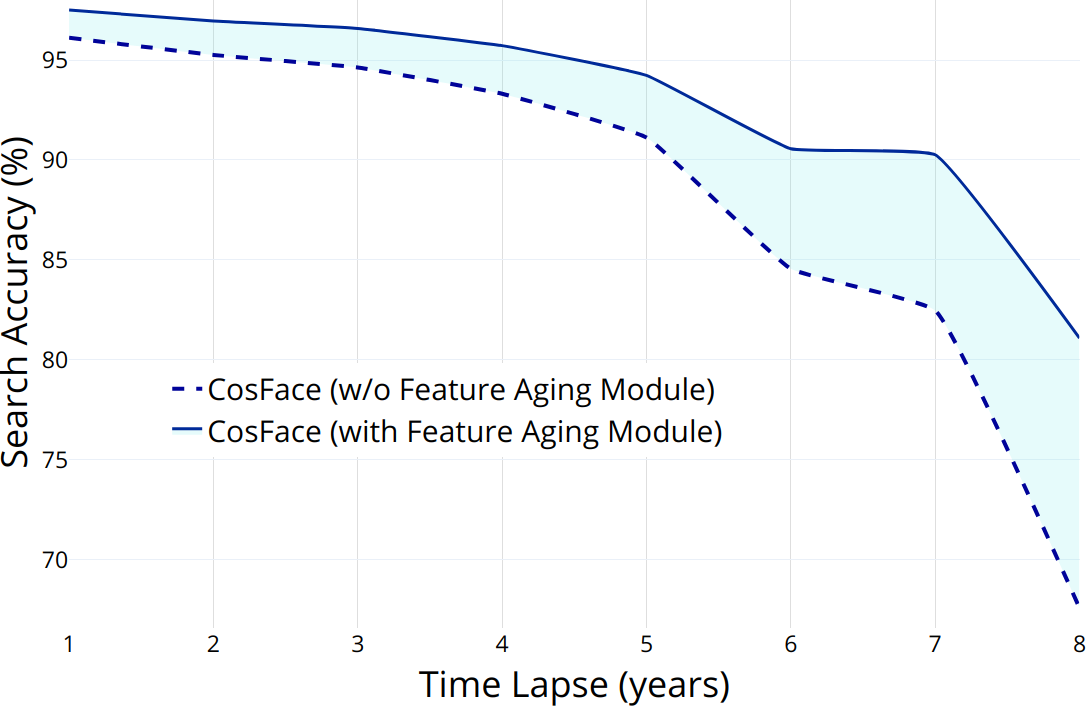}\label{fig:identification_CLF}}\hspace{1em}
    \subfloat[\scriptsize ITWCC]{\includegraphics[width=0.48\linewidth]{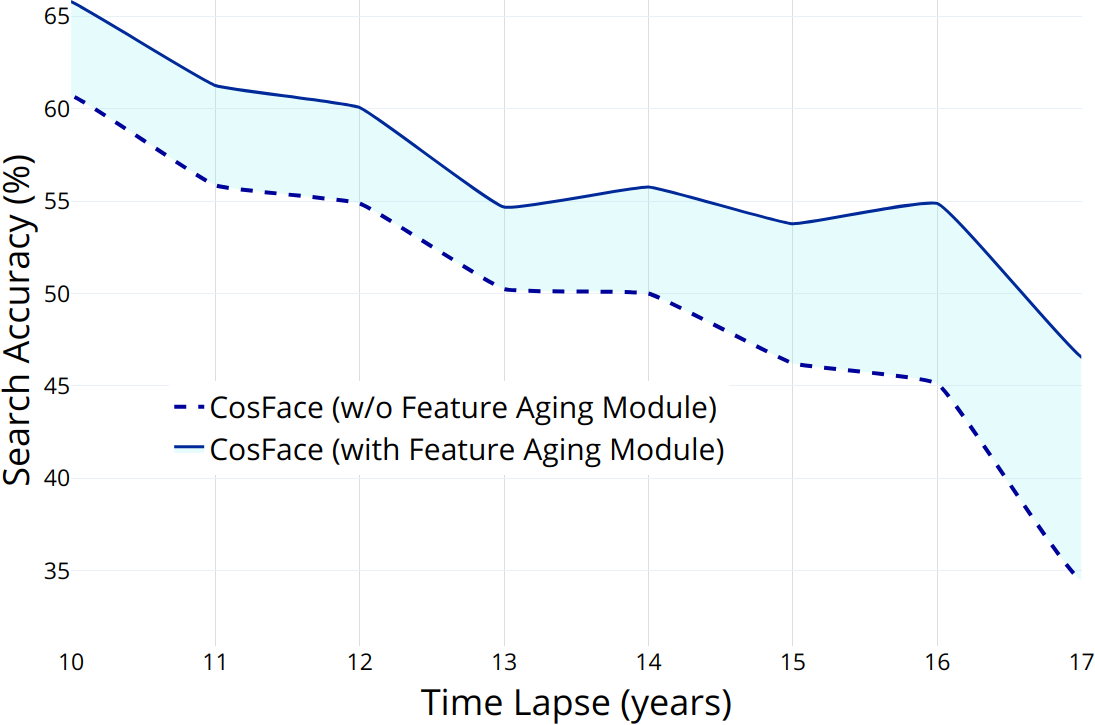}\label{fig:identification_ITWCC}}
    \caption{Rank-1 search accuracy for CosFace~\cite{cosface} on (a) CFA and (b) ITWCC datasets with and without the proposed Feature Aging Module.}
    \label{fig:identification_plots}
\end{figure*}


\paragraph{Evaluation Protocol}
We divide the children in CFA and ITWCC datasets into $2$ non-overlapping sets which are used for training and testing, respectively. Testing set of ITWCC (CFA) consists of all subjects with image pairs separated by at least a 10 (5) year time lapse. As mentioned earlier, locating missing children is akin to the identification scenario, we compute both the \emph{closed-set} identification accuracy (recovered child is in the gallery) at rank-1 and the rank-1 \emph{open-set} identification accuracy (recovered child may or may not be in the gallery) at $1.0\%$ False Accept Rate. A probe and its true mate in the gallery is separated by at least 10 years and 5 years for ITWCC and CFA, respectively. For open-set identification scenarios, we extend the gallery of CFA by adding all images in the testing of ITWCC and vice-versa. Remaining images of the subject form the distractor set in the gallery. For the search experiments, we age all the gallery features to the age of the probe and match the aged features (or aged images via our Image Generator) to the probe's face feature. 

\begin{table*}[t]
\captionsetup{font=scriptsize}
\caption{Face recognition performance on FG-NET~\cite{fgnet} and CACD-VS~\cite{cacd}.}
\vspace{0.5em}
\begin{minipage}{.5\linewidth}
\centering
\scriptsize
\begin{tabular}{l|c}
\noalign{\hrule height 0.5pt}
              \textbf{Method} & \textbf{Rank-1 (\%)}\\ \hline
\noalign{\hrule height 0.1pt}
                HFA~\cite{hfa} &  69.00\%\\
                MEFA~\cite{mefa} &  76.20\%\\
                CAN~\cite{coupled} &  86.50\%\\
                LF-CNN~\cite{lfcnn} &  88.10\%\\
                AIM~\cite{look_through_elapse} &  93.20\%\\
                Wang~\etal~\cite{decorrelated} & 94.50\%\\
                \noalign{\hrule height 0.5pt}
                COTS & 93.61\%\\
                CosFace~\cite{cosface} (w/o Feature Aging Module)&  94.91\%\\
                CosFace (finetuned on children) & 93.71\%\\
                \textbf{CosFace (with Feature Aging Module)} &  \textbf{95.91\%}\\
\noalign{\hrule height 0.5pt}
\end{tabular}\\
\medskip FG-NET\end{minipage}\hfill
\begin{minipage}{.5\linewidth}
\centering
\scriptsize
\begin{tabular}{l|c}
\noalign{\hrule height 0.5pt}
              \textbf{Method} & \textbf{Accuracy (\%)}\\ \hline
\noalign{\hrule height 0.1pt}
                High-Dimensional LBP~\cite{high_lbp} & 81.60\%\\
                HFA~\cite{hfa} &  84.40\%\\
                CARC~\cite{cacd} &  87.60\%\\
                LF-CNN~\cite{lfcnn} &  98.50\%\\
                OE-CNN~\cite{oecnn} &  99.20\%\\
                AIM~\cite{look_through_elapse} & 99.38\%\\
                Wang~\etal~\cite{decorrelated} & 99.40\%\\
                \noalign{\hrule height 0.5pt}
                COTS & 99.32\%\\
                CosFace~\cite{cosface} (w/o Feature Aging Module)&  99.50\%\\
                \textbf{CosFace (with Feature Aging Module)} &  \textbf{99.58\%}\\
\noalign{\hrule height 0.5pt}
\end{tabular}\\
\medskip CACD-VS
\end{minipage}
\label{tab:fgnet}
\end{table*}

\paragraph{Results} In Table~\ref{tab:results}, we report the Rank-1 search accuracy of our proposed Feature Aging Module (FAM) as well as accuracy when we search via our synthesized age-progressed face images on CFA and ITWCC. We find that the age-progression scheme can improve the search accuracy of both FaceNet~\cite{facenet} and CosFace~\cite{cosface}. Indeed, images aged via the FAM can also enhance the performance of these matchers which highlights the superiority of both the feature aging scheme along with the proposed Image Generator. With the proposed feature aging module, an open-source face matcher CosFace~\cite{cosface} can outperform the COTS matcher\footnote{CosFace~\cite{cosface} matcher takes about $1.56$ms to search for a probe in a gallery of $10,000$ images of missing children, while our model takes approximately $27.45$ms (on a GTX 1080 Ti) to search for a probe through the same gallery size.}.

We also investigate the Rank-1 identification rate with varying time lapses between the probe and its true mate in the gallery in  Figures~\ref{fig:identification_CLF} and~\ref{fig:identification_ITWCC}. While our aging model improves matching across all time lapses, its contribution gets larger as the time lapse increases. 

In Figure~\ref{fig:fgnet_examples}, we show some example cases where CosFace~\cite{cosface}, without the proposed deep feature aging module, retrieves a wrong child from the gallery at rank-1. With the proposed method, we can correctly identify the true mates and retrieve them at Rank-1.

In order to evaluate the generalizability of our module to adults, we train it on CFA and ITWCC~\cite{ITWCC-D1} datasets and benchmark our performance on publicly available aging datasets, FG-NET~\cite{fgnet} and CACD-VS~\cite{cacd}\footnote{\scriptsize Since CACD-VS does not have age labels, we use DEX~\cite{dex} (a publicly available age estimator) to estimate the ages.} in Table~\ref{tab:fgnet}. We follow the standard protocols~\cite{discriminative,hfa} for the two datasets and benchmark our Feature Aging Module against baselines. We find that our proposed feature aging module can enhance the performance of CosFace~\cite{cosface}. We also fine-tuned the last layer of CosFace on the same training set as ours, however, the decrease in accuracy clearly suggests that moving to a new latent space can inhibit the original features and affect general face recognition performance. Our module can boost the performance while still operating in the same feature space as the original matcher. While the two datasets do contain children, the protocol does not explicitly test large time gap between probe and gallery images of children. Therefore, we also evaluated the performance of the proposed approach on children in FG-NET with the true mate and the probe being separated by atleast 10 years. The proposed method was able to boost the performance of Cosface from 12.74\% to 15.09\%. Examples where the proposed approach is able to retrieve the true mate correctly at Rank 1 are shown in Fig.~\ref{fig:fgnet_examples}.

\begin{figure*}[!t]
\scriptsize
\centering
\includegraphics[width=0.7\linewidth]{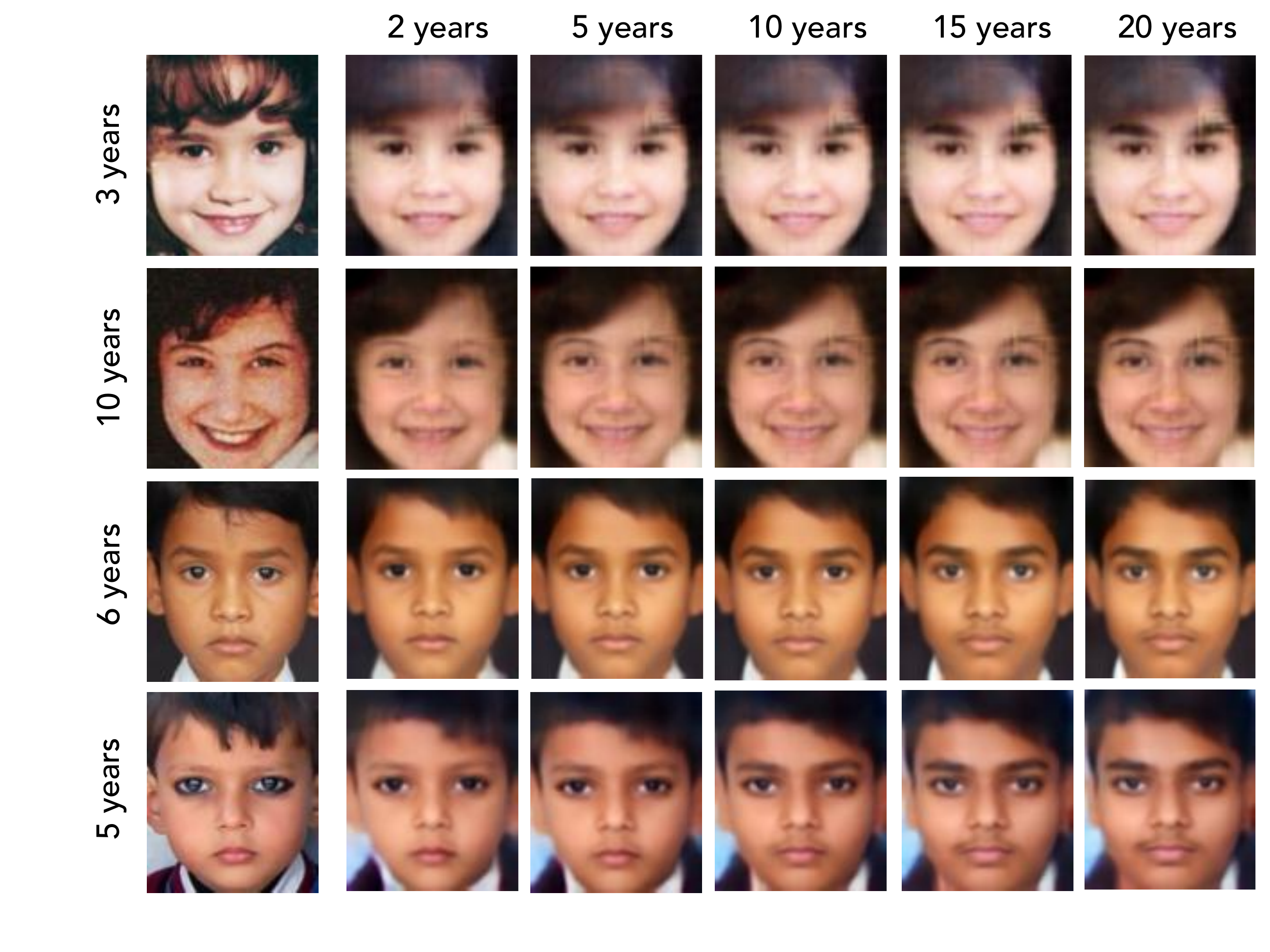}
    \caption{Column 1 shows the probe images of 4 children. Rows 1 and 2 consists of two child celebrities from the ITWCC dataset~\cite{ITWCC} and rows 3 and 4 are children from the CFA dataset. Columns 2-7 show their aged images via proposed aging scheme. The proposed aging scheme can output an aged image at any desired target age.}
     \label{fig:collage}
\end{figure*}

\subsection{Qualitative Results}
Via the proposed aging scheme, we show synthesized aged images of 4 children in the ITWCC and CFA datasets in Figure~\ref{fig:collage}. Even though we show examples when images are aged to $[2, 20]$ years with around 5 year time lapses between them, unlike prior generative methods~\cite{geng,caae,ipcgan,cgan,hongyu,lanitis}, we can synthesize an age-progressed or age-regressed image to \emph{any} desired age. We find that the proposed Feature Aging Module significantly contributes to a continuous aging process in the image space where other covariates such as poses, quality, etc. remain unchanged from the input probes. In addition, we also find that the identity of the children in the synthesized images during the aging process remains the same.

\begin{figure}[!t]
\captionsetup{font=scriptsize}
\scriptsize
\setlength{\fboxrule}{0.3em}
\centering\begin{tabular}{cccc}
& & Proposed & \\
Probe & True Mate & (Tested on FaceNet) & FaceNet~\cite{facenet}\\
19 years & 7 years & 19 years & 3 years\\
\includegraphics[width=0.12\linewidth]{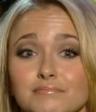} & 
\includegraphics[width=0.12\linewidth]{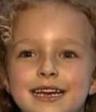} & 
\fcolorbox{Green}{White}{\includegraphics[width=0.12\linewidth]{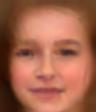}} & 
\fcolorbox{Red}{White}{\includegraphics[width=0.12\linewidth]{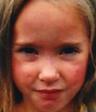}}\\
& Rank-43 & Rank-1 &%
\end{tabular}
\caption{Column 1 shows a probe image while its true mate is shown in column 2. Originally, FaceNet retrieves the true mate at Rank-43 while image in column in 4 is retrieved at Rank-1. With our aged image \emph{trained on CosFace}, FaceNet can retrieve the true mate at Rank-1.}%
\label{fig:facenet_success}
\end{figure}

\subsection{Generalization Study} Prior studies in the adversarial machine learning domain have found that deep convolutional neural networks are highly transferable, that is, different networks extract similar deep features. Since our proposed feature aging module learns to age features in the latent space of one matcher, directly utilizing the aged features for matching via another matcher may not work. However, since all matchers extract features from face images, we tested the cross-age face recognition performance on FaceNet when our module is trained via CosFace. On the ITWCC dataset~\cite{ITWCC-D1}, FaceNet originally achieves 16.53\% Rank-1 accuracy whereas, when we test the aged images, FaceNet achieves \textbf{21.11\%} Rank-1 identification rate. We show an example where the proposed scheme trained on CosFace can aid FaceNet's longitudinal face recognition performance in Figure~\ref{fig:facenet_success}.

\subsection{Ablation Study} 
\begin{figure*}[!t]
    \centering
    \scriptsize
    \captionsetup{font=scriptsize}
    \begin{tabular}{ccccc}
    Original Image & w/o $\EE_{style}$ & w/o $\EL_{pix}$ & w/o $\EL_{ID}$ & with all\\
    \includegraphics[height=0.6in]{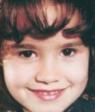} &
    \includegraphics[height=0.6in]{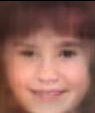} &
    \includegraphics[height=0.6in]{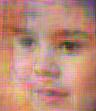} & \includegraphics[height=0.6in]{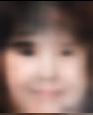} &
     \includegraphics[height=0.6in]{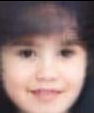}\\
     & \textbf{0.66} & \textbf{0.83}  & \textbf{0.34} & \textbf{0.74}
    \end{tabular}
    \caption{A real face image along with variants of the proposed Image Generator trained without style-encoder, pixel-supervision loss, and identity loss, respectively. Last column shows an example of the proposed method comprised of all. Denoted below each image is the cosine similarity score (from CosFace) to the original image.}
     \label{fig:ablation_1}
\end{figure*}

In order to analyze the importance of each loss function in the final objective function for the Image Generator, in Figure~\ref{fig:ablation_1}, we train three variants of the generator: (1) removed the style-encoder ($\EE_{style}$), (2) without pixel-level supervision ($\EL_{pix}$), and (3) without identity-preservation ($\EL_{ID}$), respectively.  The style-encoder helps to maintain the visual quality
of the synthesized faces. With the decoder alone, undesirable artifacts are introduced due to no style information from the image. Without pixel-level supervision, generative noises are introduced which leads to a lack in perceptual quality. This is because the decoder is instructed to construct \emph{any} image that can match in the latent space of the matcher, which may not align with the perceived visual quality of the image. The identity loss is imperative in ensuring that the synthesized image can be used for face recognition. Without the identity loss, the
synthesized image cannot aid in cross-age face recognition. We find that every component of the proposed objective function is necessary
in order to obtain a synthesized face that is not only visually appealing but can also maintain the identity of the probe.

\subsection{Discussion}
\begin{figure*}[!t]
    \centering
    \captionsetup{font=scriptsize}
    \subfloat[]{\includegraphics[height=2.1in]{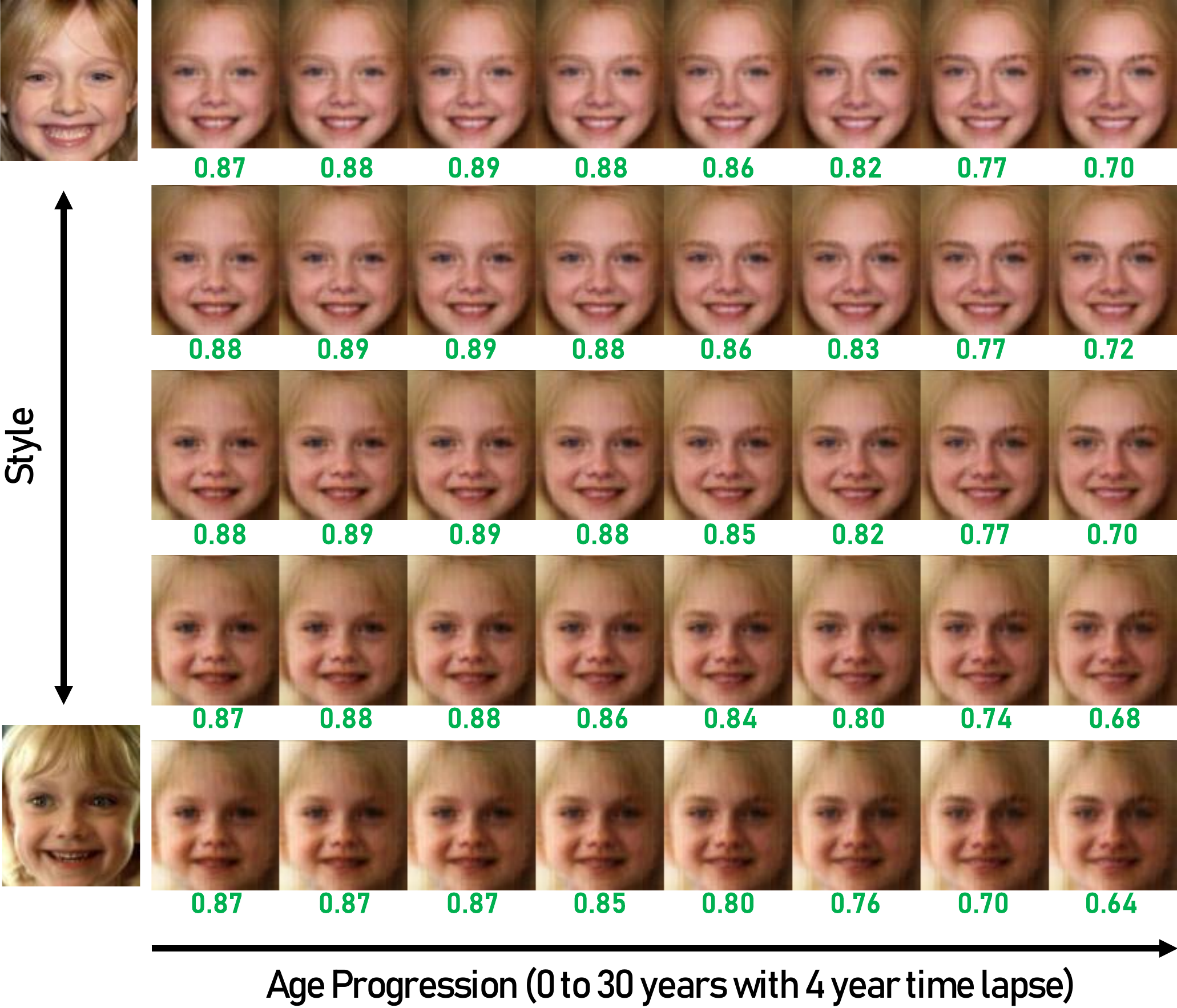}}\hfill
    \subfloat[]{\includegraphics[height=2.1in]{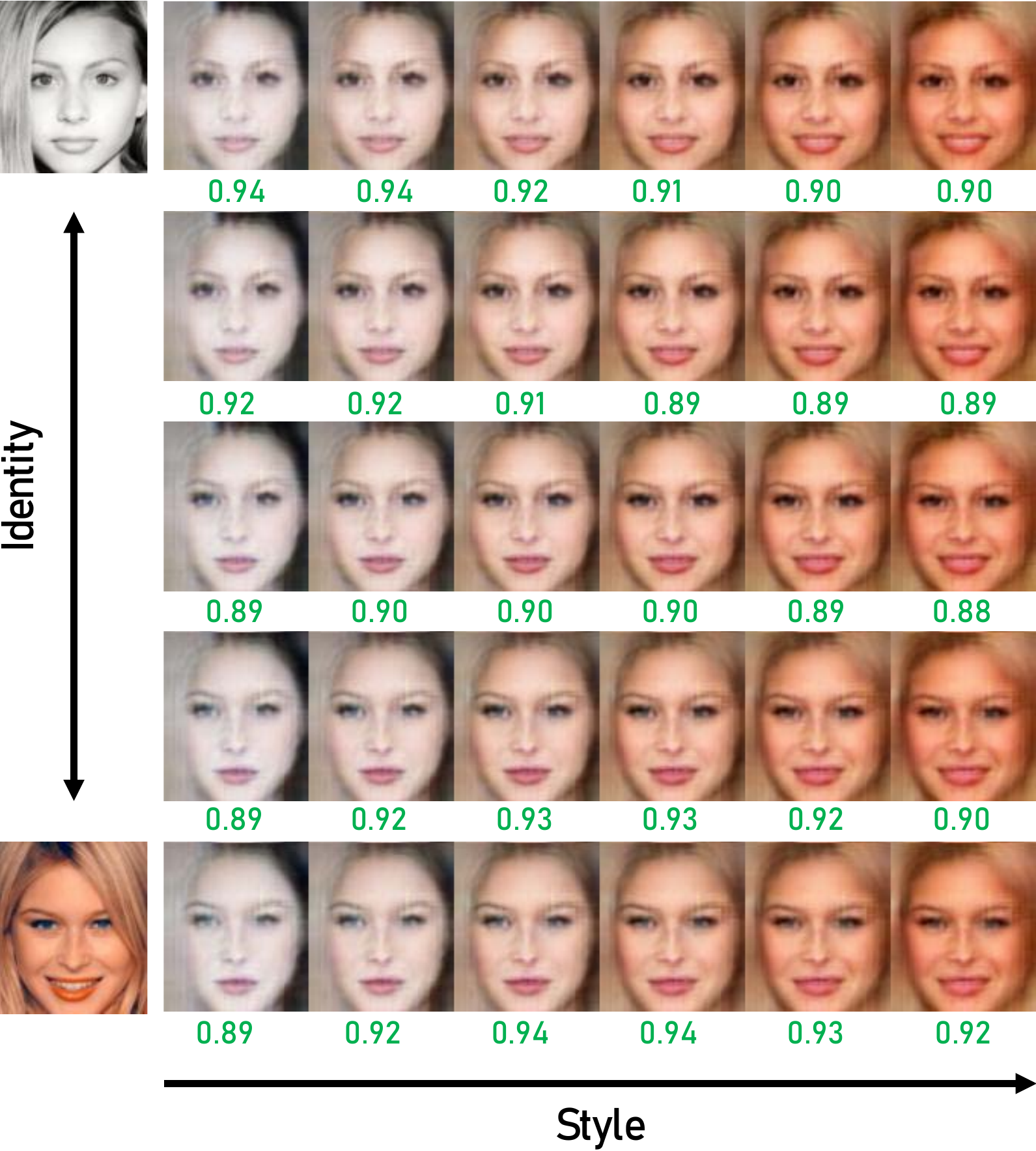}}
    \caption{Interpolation between style, age, and identity. (a) Shows that aged features by the proposed FAM does not alter other covariates such as quality and pose, while the style transitions due to the style-encoder. (b) Shows the gradual transitioning between style and identity which verifies the effictiveness of our method in disentangling identity and non-identity (style) features.}
    \label{fig:interpolation}
\end{figure*}

\paragraph{Interpolation}
From our framework, we can obtain the style vector from the style-encoder and the identity feature from the identity encoder. In Figure~\ref{fig:interpolation}(a), in the y-axis, we interpolate between the styles of the two images, both belonging to the same child and acquired at age 7. In the x-axis, we synthesize the aged images by passing the interpolated style feature and the aged feature from the Feature Aging Module. We also provide the cosine similarity scores obtained from CosFace when the aged images are compared to the youngest synthesized image. We show that throughout the aging process, the proposed FAM and Image Generator can both preserve the identity of the child. In addition, it can be seen that the FAM module does not affect other covariates such as pose or quality.
In Figure~\ref{fig:interpolation}(b), we take two images of two different individuals with vastly different styles. We then interpolate between the style in the x-axis and the identity in the y-axis. We see a smooth transitions between style and identity which verifies that the proposed style-encoder and decoder are effective in disentangling identity and non-identity (style) features.

\subsection{Two Case Studies of Missing Children}\label{sec:case_study}
Carlina White was abducted from the Harlem hospital center in New York City when she was just $19$ days old. She was reunited with her parents $23$ years later when she saw a photo resembling her as a baby on the National Center for Missing and Exploited Children website\footnote{\scriptsize    \url{http://www.missingkids.org}}. We constructed a gallery of missing children consisting of $12,873$ face images in the age range $0$ - $32$ years from the UTKFace~\cite{caae} dataset and Carlina's image as an infant when she went missing (19 days old). Her face image when she was later found (23 years old) was used as the probe.
State-of-the-art face matchers, CosFace~\cite{cosface} and COTS, were able to retrieve probe's true mate at ranks $3,069$ and $1,242$ respectively. It is infeasible for a human operator to look through such a large number of retrieved images to ascertain the true mate. With the proposed FAM, CosFace is able to retrieve the true mate at \textbf{Rank-268}, which is a significant improvement in narrowing down the search.

\begin{figure}[!t]
\scriptsize
\centering
\captionsetup{font=scriptsize}
\setlength{\fboxrule}{0.2em}
\begin{tabular}{c@{ }c@{ }c@{ }}
\scriptsize
\textbf{Probe} & \textbf{Gallery} &  \textbf{Age Progressed}\\
11 years & 0 years & 13 years \\
\includegraphics[height=0.6in]{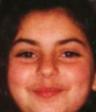} & 
\includegraphics[height=0.6in]{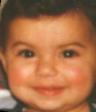} &
\includegraphics[height=0.6in]{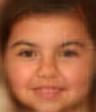}\\[-0.2em]
& \textbf{\textcolor{Red}{Rank-535}} & \textbf{\textcolor{Green}{Rank-1}} \\[0.5em]
17 years & 4 years & 19 years\\
\includegraphics[height=0.6in]{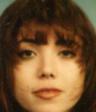}&
\includegraphics[height=0.6in]{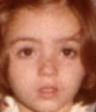} &
\includegraphics[height=0.6in]{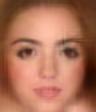}\\[-0.2em]
& \textbf{\textcolor{Red}{Rank-187}} & \textbf{\textcolor{Green}{Rank-1}}
\end{tabular}\hfill
\begin{tabular}{@{}c@{ }c@{ }c@{ }}
\scriptsize
\textbf{Probe} & \textbf{CosFace}~\cite{cosface} & \textbf{CosFace + Aging}\\[0.2em]
\includegraphics[height=0.6in]{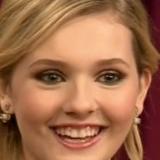}&
\fcolorbox{Red}{white}{\includegraphics[height=0.6in]{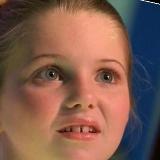}}&
\fcolorbox{Green}{white}{\includegraphics[height=0.6in]{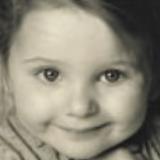}}\\
15 years & 8 years & 2 years\\
\includegraphics[height=0.6in]{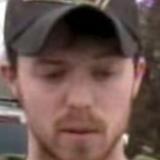}&
\fcolorbox{Red}{white}{\includegraphics[height=0.6in]{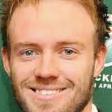}}&
\fcolorbox{Green}{white}{\includegraphics[height=0.6in]{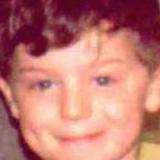}}\\
24 years & 28 years & 5 years
\end{tabular}
\caption{Identities incorrectly retrieved at Rank-1 by CosFace~\cite{cosface} without our proposed age-progression module (highlighted in red). CosFace with the proposed method can correctly retrieve the true mates in the gallery at rank-1 (highlighted in green).}%
\label{fig:fgnet_examples}
\end{figure}

In another missing child case, Richard Wayne Landers was abducted by his grandparents at age $5$ in July 1994 in Indiana. In 2013, investigators identified Richard (then, $24$ years old) through a Social Security database search (see Figure~\ref{fig:fgnet_examples}). Akin to Carlina's case, adding Richard's $5$ year old face image in the gallery and keeping his face image at age $24$ as the probe, CosFace~\cite{cosface} was able to retrieve his younger image at Rank-23. With the proposed feature aging, CosFace was able to retrieve his younger image at \textbf{Rank-1}.


These examples show the applicability of our feature aging module to real world missing children cases. By improving the search accuracy of any face matcher in children-to-adult matching, our model makes a significant contribution to the social good by reuniting missing children with their loved ones.

\section{Summary}
We propose a new method that can age deep face features in order to enhance the longitudinal face recognition performance in identifying missing children. The proposed method also guides the aging process in the image space such that the synthesized aged images can boost cross-age face recognition accuracy of any commodity face matcher. The proposed approach enhances the rank-1 identification accuracies of FaceNet from 16.53\% to 21.44\% and CosFace from 60.72\% to 66.12\% on a child celebrity dataset, namely, ITWCC. Moreover, with the proposed method, rank-1 accuracy of CosFace on a public aging face dataset, FG-NET, increases from 94.91\% to 95.91\%, outperforming state-of-the-art. These results suggest that the proposed aging scheme can enhance the ability of commodity face matchers to locate and identify young children who are lost at a young age in order to reunite them back with their families. We plan to extend our work to unconstrained child face images which is typical in child trafficking cases.

\bibliographystyle{splncs04}
\bibliography{egbib}

\begin{thebibliography}{10}
\providecommand{\url}[1]{\texttt{#1}}
\providecommand{\urlprefix}{URL }
\providecommand{\doi}[1]{https://doi.org/#1}

\bibitem{Child}
{Convention on the Rights of the Child}.
  \url{https://web.archive.org/web/20101031104336/http://www.hakani.org/en/convention/Convention_Rights_Child.pdf}
  (1989)

\bibitem{sherbat}
{A life revealed}. \url{http://www. nationalgeographic.com/magazine/2002/
  04/afghan-girl-revealed/} (2002)

\bibitem{fgnet}
{FG-NET dataset}. \url{https://yanweifu.github.io/FG_NET_data/index.html}
  (2014)

\bibitem{UNICEF}
{Children account for nearly one-third of identified trafficking victims
  globally}. \url{https://uni.cf/2OqsMIt} (2018)

\bibitem{cgan}
Antipov, G., Baccouche, M., Dugelay, J.L.: Face aging with conditional
  generative adversarial networks. In: IEEE ICIP (2017)

\bibitem{bereta}
Bereta, M., Karczmarek, P., Pedrycz, W., Reformat, M.: Local descriptors in
  application to the aging problem in face recognition. Pattern Recognition
  \textbf{46}(10),  2634--2646 (2013)

\bibitem{lacey}
Best-Rowden, L., Hoole, Y., Jain, A.: Automatic face recognition of newborns,
  infants, and toddlers: A longitudinal evaluation. In: {IEEE BIOSIG}. pp.~1--8
  (2016)

\bibitem{lacey_adult}
Best-Rowden, L., Jain, A.K.: Longitudinal study of automatic face recognition.
  IEEE T-PAMI  \textbf{40}(1),  148--162 (2017)

\bibitem{vggface2}
Cao, Q., Shen, L., Xie, W., Parkhi, O.M., Zisserman, A.: Vggface2: A dataset
  for recognising faces across pose and age. In: IEEE FG (2018)

\bibitem{cacd}
Chen, B.C., Chen, C.S., Hsu, W.H.: Cross-age reference coding for age-invariant
  face recognition and retrieval. In: ECCV (2014)

\bibitem{high_lbp}
Chen, D., Cao, X., Wen, F., Sun, J.: Blessing of dimensionality:
  High-dimensional feature and its efficient compression for face verification.
  In: CVPR (2013)

\bibitem{anatomy_face}
Coleman, S.R., Grover, R.: {The Anatomy of the Aging Face: Volume Loss and
  Changes in 3-Dimensional Topography}. Aesthetic Surgery Journal  \textbf{26},
   S4--S9 (2006). \doi{10.1016/j.asj.2005.09.012},
  \url{https://doi.org/10.1016/j.asj.2005.09.012}

\bibitem{saroo_movie}
Davis, G.: \url{http://lionmovie.com/} (2016)

\bibitem{deb_adult}
Deb, D., Best-Rowden, L., Jain, A.K.: Face recognition performance under aging.
  In: CVPRW (2017)

\bibitem{deb_child}
Deb, D., Nain, N., Jain, A.K.: Longitudinal study of child face recognition.
  In: IEEE ICB (2018)

\bibitem{arcface}
Deng, J., Guo, J., Xue, N., Zafeiriou, S.: Arcface: Additive angular margin
  loss for deep face recognition. In: CVPR (2019)

\bibitem{attributes}
Dhar, P., Bansal, A., Castillo, C.D., Gleason, J., Phillips, P.J., Chellappa,
  R.: How are attributes expressed in face dcnns? arXiv preprint
  arXiv:1910.05657  (2019)

\bibitem{geng}
Geng, X., Zhou, Z.H., Smith-Miles, K.: Automatic age estimation based on facial
  aging patterns. IEEE T-PAMI  \textbf{29}(12),  2234--2240 (2007)

\bibitem{hfa}
Gong, D., Li, Z., Lin, D., Liu, J., Tang, X.: Hidden factor analysis for age
  invariant face recognition. In: CVPR (2013)

\bibitem{mefa}
Gong, D., Li, Z., Tao, D., Liu, J., Li, X.: A maximum entropy feature
  descriptor for age invariant face recognition. In: CVPR (2015)

\bibitem{nist_irex_report}
Grother, P.J., Matey, J.R., Tabassi, E., Quinn, G.W., Chumakov, M.: Irex
  vi-temporal stability of iris recognition accuracy. {NIST Interagency Report}
   \textbf{7948} (2013)

\bibitem{nist_2018}
Grother, P.J., Ngan, M., Hanaoka, K.: {Ongoing Face Recognition Vendor Test
  (FRVT), Part 2: Identification}. {NIST Interagency Report}  (2018)

\bibitem{nist}
Grother, P.J., Quinn, G.W., Phillips, P.J.: Report on the evaluation of 2d
  still-image face recognition algorithms. {NIST Interagency Report}
  \textbf{7709}, ~106 (2010)

\bibitem{hill}
Hill, M.Q., Parde, C.J., Castillo, C.D., Colon, Y.I., Ranjan, R., Chen, J.C.,
  Blanz, V., O'Toole, A.J.: Deep convolutional neural networks in the face of
  caricature: Identity and image revealed. arXiv preprint arXiv:1812.10902
  (2018)

\bibitem{klare}
Klare, B., Jain, A.K.: Face recognition across time lapse: On learning feature
  subspaces. In: IEEE IJCB (2011)

\bibitem{lanitis}
Lanitis, A., Taylor, C.J., Cootes, T.F.: Toward automatic simulation of aging
  effects on face images. IEEE T-PAMI  \textbf{24}(4),  442--455 (2002)

\bibitem{hierarchical}
Li, Z., Gong, D., Li, X., Tao, D.: Aging face recognition: A hierarchical
  learning model based on local patterns selection. IEEE TIP  \textbf{25}(5),
  2146--2154 (2016)

\bibitem{discriminative}
Li, Z., Park, U., Jain, A.K.: A discriminative model for age invariant face
  recognition. IEEE TIFS  \textbf{6}(3),  1028--1037 (2011)

\bibitem{ling}
Ling, H., Soatto, S., Ramanathan, N., Jacobs, D.W.: Face verification across
  age progression using discriminative methods. IEEE TIFS  \textbf{5}(1),
  82--91 (2009)

\bibitem{sphereface}
Liu, W., Wen, Y., Yu, Z., Li, M., Raj, B., Song, L.: Sphereface: Deep
  hypersphere embedding for face recognition. In: CVPR (2017)

\bibitem{lfcnn}
Nhan~Duong, C., Gia~Quach, K., Luu, K., Le, N., Savvides, M.: Temporal
  non-volume preserving approach to facial age-progression and age-invariant
  face recognition. In: ICCV (2017)

\bibitem{unsang}
Park, U., Tong, Y., Jain, A.K.: Age-invariant face recognition. IEEE
  Transactions on Pattern Analysis and Machine Intelligence  \textbf{32}(5),
  947--954 (2010)

\bibitem{facialstructure}
Ramanathan, N., Chellappa, R.: Modeling age progression in young faces. In:
  CVPR (2006). \doi{10.1109/CVPR.2006.187}

\bibitem{ITWCC}
Ricanek, K., Bhardwaj, S., Sodomsky, M.: {A review of face recognition against
  longitudinal child faces}. In: IEEE BIOSIG (2015)

\bibitem{morph}
Ricanek, K., Tesafaye, T.: Morph: A longitudinal image database of normal adult
  age-progression. In: IEEE FG (2006)

\bibitem{dex}
Rothe, R., Timofte, R., Van~Gool, L.: Dex: Deep expectation of apparent age
  from a single image. In: ICCV Workshop

\bibitem{facenet}
Schroff, F., Kalenichenko, D., Philbin, J.: Facenet: A unified embedding for
  face recognition and clustering. In: CVPR. pp. 815--823 (2015)

\bibitem{ITWCC-D1}
Srinivas, N., Ricanek, K., Michalski, D., Bolme, D.S., King, M.A.: {Face
  Recognition Algorithm Bias: Performance Differences on Images of Children and
  Adults}. In: CVPR Workshops (2019)

\bibitem{decorrelated}
Wang, H., Gong, D., Li, Z., Liu, W.: Decorrelated adversarial learning for
  age-invariant face recognition. In: CVPR (2019)

\bibitem{cosface}
Wang, H., Wang, Y., Zhou, Z., Ji, X., Gong, D., Zhou, J., Li, Z., Liu, W.:
  Cosface: Large margin cosine loss for deep face recognition. In: CVPR (2018)

\bibitem{recurrent}
Wang, W., Cui, Z., Yan, Y., Feng, J., Yan, S., Shu, X., Sebe, N.: {Recurrent
  Face Aging}. In: CVPR (2016). \doi{10.1109/CVPR.2016.261}

\bibitem{oecnn}
Wang, Y., Gong, D., Zhou, Z., Ji, X., Wang, H., Li, Z., Liu, W., Zhang, T.:
  Orthogonal deep features decomposition for age-invariant face recognition.
  In: ECCV (2018)

\bibitem{coupled}
Xu, C., Liu, Q., Ye, M.: Age invariant face recognition and retrieval by
  coupled auto-encoder networks. Neurocomputing  \textbf{222},  62--71 (2017)

\bibitem{hongyu}
Yang, H., Huang, D., Wang, Y., Jain, A.K.: Learning face age progression: A
  pyramid architecture of gans. In: CVPR (2018)

\bibitem{yoon}
Yoon, S., Jain, A.K.: Longitudinal study of fingerprint recognition. {PNAS}
  \textbf{112}(28),  8555--8560 (2015)

\bibitem{ipcgan}
Zhai, Z., Zhai, J.: Identity-preserving conditional generative adversarial
  network. In: IEEE IJCNN (2018)

\bibitem{mtcnn}
Zhang, K., Zhang, Z., Li, Z., Qiao, Y.: Joint face detection and alignment
  using multitask cascaded convolutional networks. IEEE SPL  \textbf{23}(10),
  1499--1503 (2016)

\bibitem{caae}
Zhang, Z., Song, Y., Qi, H.: Age progression/regression by conditional
  adversarial autoencoder. In: CVPR (2017)

\bibitem{look_through_elapse}
Zhao, J., Cheng, Y., Cheng, Y., Yang, Y., Zhao, F., Li, J., Liu, H., Yan, S.,
  Feng, J.: {Look across elapse: Disentangled representation learning and
  photorealistic cross-age face synthesis for age-invariant face recognition}.
  In: AAAI (2019)

\bibitem{aecnn}
Zheng, T., Deng, W., Hu, J.: Age estimation guided convolutional neural network
  for age-invariant face recognition. In: CVPRW (2017)

\end{thebibliography}

\title{Appendix}
\titlerunning{} 
\authorrunning{} 
\author{}
\institute{}
\appendix
\maketitle
\section{What is Feature Aging Module Learning?}

\subsection{Age-Sensitive Features} In order to analyze which components of the face embeddings are altered during the aging process, we first consider a decoder that takes an input a face embedding and attempts to construct a face image without any supervision from the image itself. That is, the decoder is a variant of the proposed Image Generator without the style encoder. This ensures that the decoder can only synthesize a face image from whatever is encoded in the face feature only.

\begin{figure}[!h]
    \centering
    \begin{tabular}{ccccc}
         \includegraphics[width=0.18\textwidth]{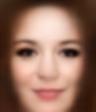}& \includegraphics[width=0.18\textwidth]{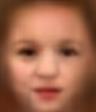}&
         \includegraphics[width=0.18\textwidth]{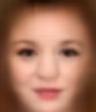}& \includegraphics[width=0.18\textwidth]{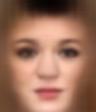}& \includegraphics[width=0.18\textwidth]{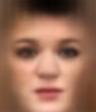}\\
         Probe (18 years) & 2 years & 10 years & 25 years & 30 years\\
        &
        \includegraphics[width=0.18\textwidth]{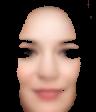}&
         \includegraphics[width=0.18\textwidth]{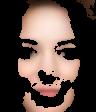}& \includegraphics[width=0.18\textwidth]{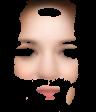}& \includegraphics[width=0.18\textwidth]{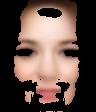}\\
         
         \includegraphics[width=0.18\textwidth]{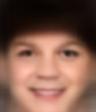}& \includegraphics[width=0.18\textwidth]{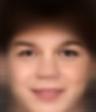}&
         \includegraphics[width=0.18\textwidth]{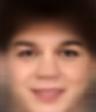}& \includegraphics[width=0.18\textwidth]{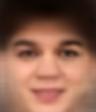}& \includegraphics[width=0.18\textwidth]{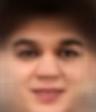}\\
         Probe (5 years) & 10 years & 15 years & 20 years & 25 years\\
        &
        \includegraphics[width=0.18\textwidth]{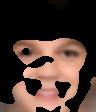}&
         \includegraphics[width=0.18\textwidth]{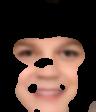}& \includegraphics[width=0.18\textwidth]{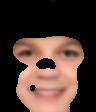}& \includegraphics[width=0.18\textwidth]{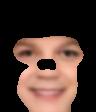}\\
    \end{tabular}
    \caption{Differences between age-progressed and age-regressed features. Rows 1 and 3 show decoded aged images for two different individuals. Rows 2 and 4 show the areas that change from the probe (black color indicates no-changes). For both progression and regression, our proposed Feature Aging Module only alters face components responsible for aging while largely keeping other covariates such as background, pose, quality, and style consistent.}
    \label{fig:visualize_direct}
\end{figure}

We then compute a difference image between a reconstructed face image and its age-progressed version. We directly age the input feature vector via our Feature Aging Module as shown in Figure~\ref{fig:visualize_direct}. We find that when a probe feature is progressed to a younger age, our method attempts to reduce the size of the head and the eyes, whereas, age-progression enlarges the head, adds makeup, and adds aging effects such as wrinkles around the cheeks. As we expect, only components responsible for aging are altered, whereas, noise factors such as background, pose, quality, and style remain consistent.

\subsection{Why does Deep Feature Aging enhance longitudinal performance?}
\begin{figure}[!h]
    \centering
    \setlength{\fboxrule}{0.3em}
    \begin{tabular}{cccccc}
        \includegraphics[width=0.15\textwidth]{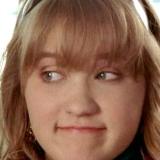} & 
        \includegraphics[width=0.15\textwidth]{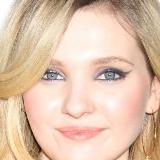} & 
        \includegraphics[width=0.15\textwidth]{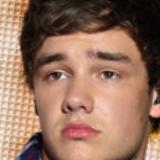} & 
        \includegraphics[width=0.15\textwidth]{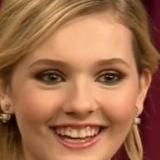} & 
        \includegraphics[width=0.15\textwidth]{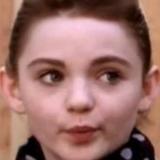} & 
        \fcolorbox{Red}{White}{\includegraphics[width=0.15\textwidth]{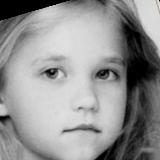}}\\
        Probe & Rank 1 & Rank 2 & Rank 3 & Rank 4 & Rank 37\\
        17 years & 15 years & 18 years & 15 years & 13 years & 5 years\\\\
        \includegraphics[width=0.15\textwidth]{new_images/longitudinal_perf/17_1_CS0007011f17.jpg}
        &
        \fcolorbox{Green}{White}{\includegraphics[width=0.15\textwidth]{new_images/longitudinal_perf/05_1_CS0007004f05.jpg}} & 
        \includegraphics[width=0.15\textwidth]{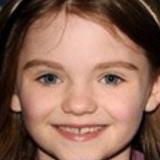} & 
        \includegraphics[width=0.15\textwidth]{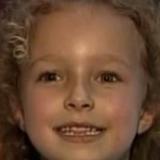} & 
        \includegraphics[width=0.15\textwidth]{new_images/longitudinal_perf/15_1_CS0027003f15.jpg} & 
        \includegraphics[width=0.15\textwidth]{new_images/longitudinal_perf/13_1_CS0094059f13.jpg}\\
         Probe & Rank 1 & Rank 2 & Rank 3 & Rank 4 & Rank 5\\
         17 years & 5 years & 7 years & 7 years & 15 years & 13 years\\
    \end{tabular}
    \caption{Row 1: CosFace wrongly retrieves the true mate at Rank-37. Top-5 retrievals include gallery images that are of similar ages as that of the probe. Row 2: With the proposed Feature Aging Module, CosFace focuses only on identity-salient features that are age-invariant and retrieves children in the top-5 retrievals. In this scenario, our Feature Aging Module can aid CosFace in retrieving the true mate at Rank-1.}
    \label{fig:longitudinal}
\end{figure}

In Figure~\ref{fig:longitudinal}, in the first row, we see that originally a state-of-the-matcher, CosFace~\cite{cosface}, wrongly retrieves the true mate at Rank-37. Interestingly, we find that the top 5 retrievals are very similar ages to the probe's age (17 years). That is, state-of-the-art matchers are biased towards retrieving images from the gallery that are of similar ages as that of the probe. With our Feature Aging Module (FAM), we age the feature in the feature space of the matcher such that we can `fool' the matcher into thinking that the gallery is closer to the probe's age. In this manner, the matcher tends to utilize identity-salient features that are age-invariant and can only focus on those facial components. In row 2, we find that when we age the gallery to the probe's age, the top retrievals are all children. This highlights the strength of our Feature Aging Module and its ability to enhance longitudinal performance of matchers.

\subsection{Effect of Deep Feature Aging on Embeddings}
\begin{figure}[!h]
\footnotesize
\setlength{\fboxrule}{0.2em}
\includegraphics[width=0.21\linewidth]{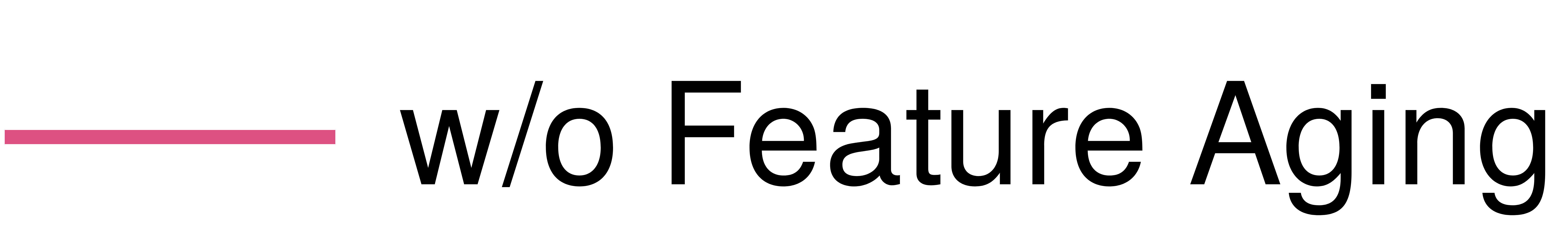}\hspace{10em}
\includegraphics[width=0.21\linewidth]{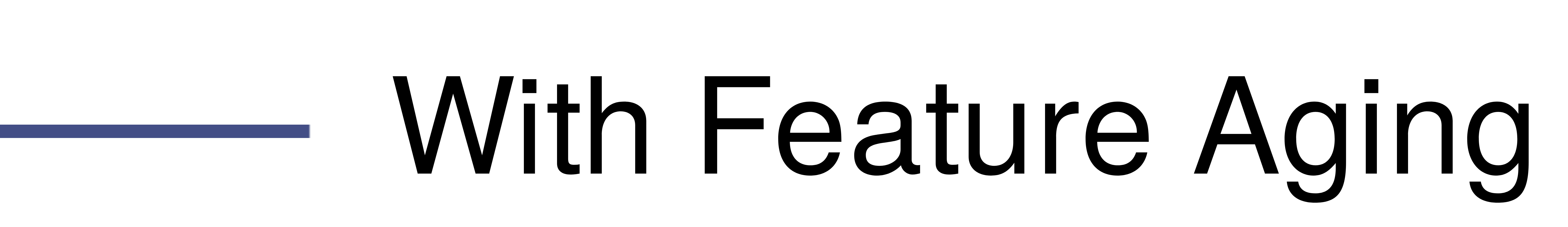}
\centering\begin{tabular}{@{}c@{ }c@{ }c@{ }c@{ }c@{ }}
\includegraphics[width=0.45\linewidth]{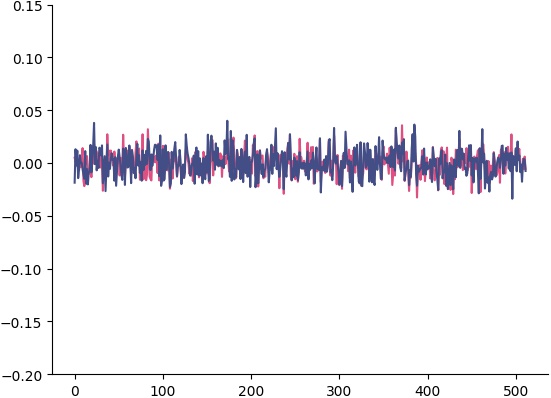} & \includegraphics[width=0.45\linewidth]{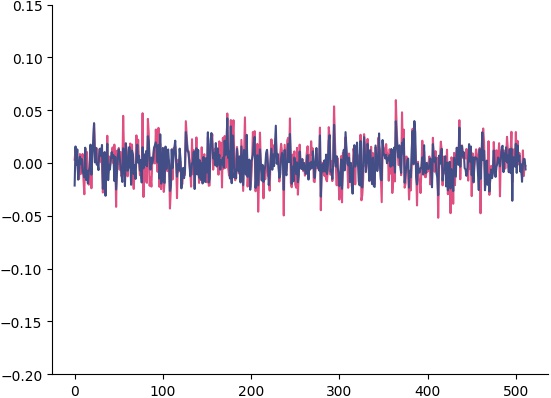} \\\vspace{-0.3em}
\textbf{\large 5 years} & \textbf{\large 7 years}\\\\
\includegraphics[width=0.45\linewidth]{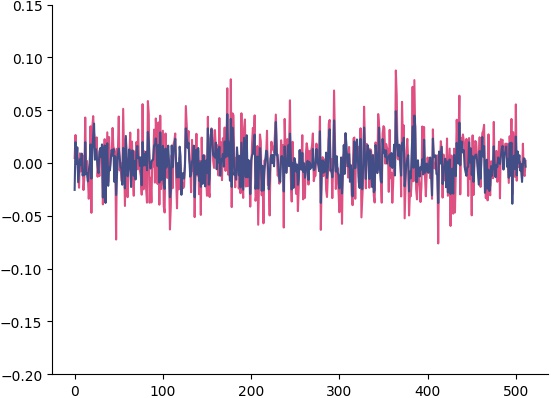} &
\includegraphics[width=0.45\linewidth]{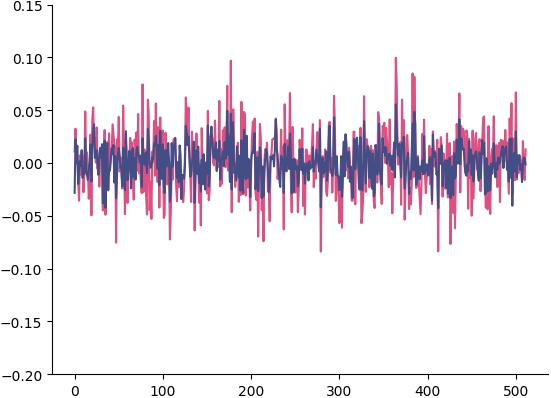}\\\vspace{-0.3em}
\textbf{\large 10 years} & \textbf{\large 12 years}\\\\
\includegraphics[width=0.45\linewidth]{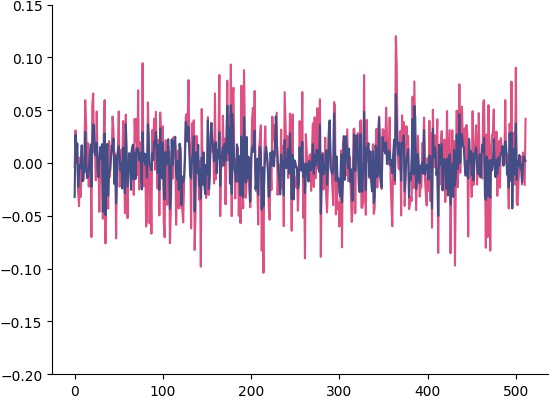} &
\includegraphics[width=0.45\linewidth]{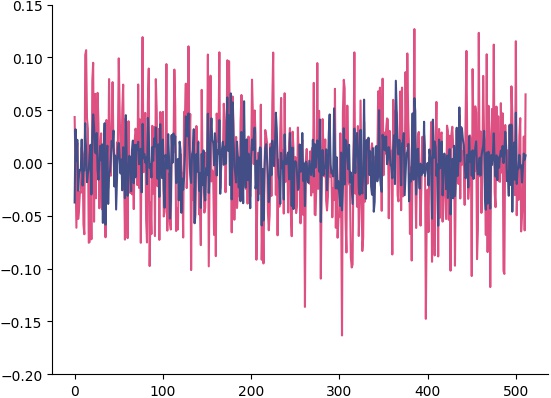}\\\vspace{-0.3em}
\textbf{\large 15 years} & \textbf{\large 20 years}
\end{tabular}
\caption{{Difference between mean feature at age 2 years and different target ages (denoted below each plot) with CosFace\cite{cosface} and CosFace\cite{cosface} with the proposed feature aging module on the CFA dataset. The difference between the mean features increases as the time lapse becomes larger but with the proposed feature aging module the difference is relatively constant over time lapse. This clearly shows the superiority of our module over the original matcher.}}
\label{fig:mean_feature}
\end{figure}

To observe the effect of our module on the face embeddings, we plot the difference between the mean feature vectors of all subjects (in the test set) at the youngest age in the CFA dataset, \ie~2 years, and mean feature vectors at different target ages (in the test set) (see Figure~\ref{fig:mean_feature}).

For a state-of-the-art face matcher, CosFace\cite{cosface}, the differences between these mean feature vectors, over all 512 dimensions, increases over time lapse causing the recognition accuracy of the matcher to drop for larger time lapses. However, with the proposed feature aging module, the difference remains relatively constant as the time lapse increases. This indicates that the proposed feature aging module is able to maintain relatively similar performance over time lapses.

\section{Effect of Style Dimensionality}
In this section, we evaluate the effect of increasing or decreasing the dimensionality of the style vector obtained via the proposed Style Encoder. Given a style vector of $k$-dimensions, we concatenate the style vector to a $d$-dimensional ID feature vector extracted via an ID encoder. For this experiment, we consider a $512$-dimensional ID embedding obtained via CosFace~\cite{cosface}.

\begin{figure}[!h]
\centering
    \includegraphics[width=0.6\linewidth]{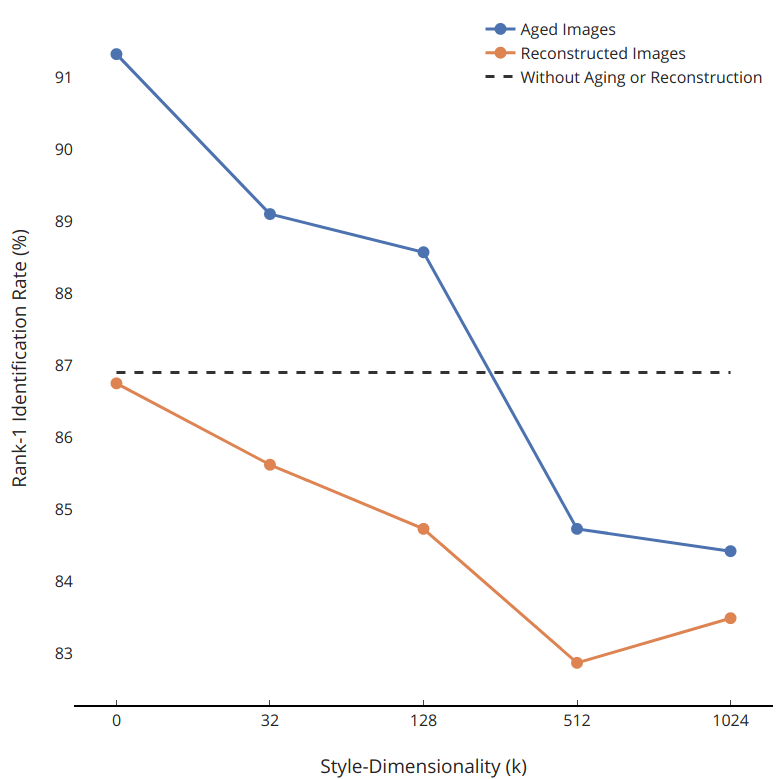}
    \caption{Trade-off between identification rates from reconstruction and aging. For our experiments, we choose $k=32$.}
    \label{fig:style_dim}
\end{figure}

\begin{figure}[!h]
\centering
    \includegraphics[width=\linewidth]{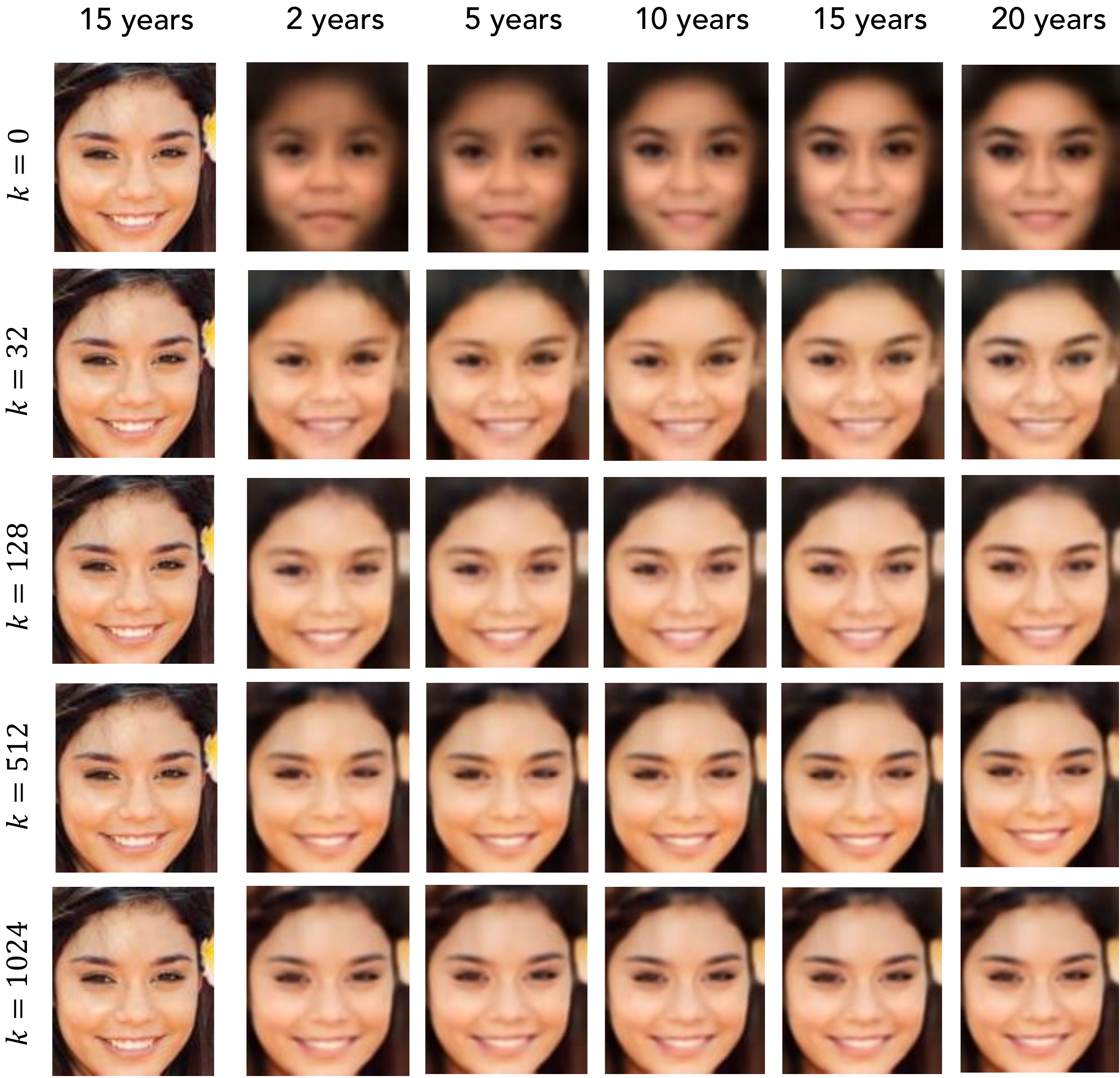}
    \caption{Trade-off between visual quality and aging. For our experiments, we choose $k=32$.}
    \label{fig:style_dim_fig}
\end{figure}

We evaluate the identification rate when the gallery is aged to the probe's age as well simply reconstructing the gallery to its own age. We train our proposed framework on ITWCC training dataset~\cite{ITWCC-D1} and evaluate the identification rate on a validation set of CLF dataset. Note that we never conduct any experiment on the testing sets of either ITWCC nor CLF datasets. In Figure~\ref{fig:style_dim}, we observe a clear trade-off between reconstruction accuracy and aging accuracy for $k=0,~32,~128,~512,~1024$. That is, for larger values of $k$, the decoder tends to utilize more style-specific information while ignoring the ID feature (which is responsible for aging via FAM). In addition, the gap between reconstruction and aging accuracy narrows as $k$ gets larger due to the decoder ignoring the identity feature vector from the ID encoder. In Figure~\ref{fig:style_dim_fig}, we can observe this trade-off clearly. Larger $k$ enforces better perceptual quality among the synthesized images, with lesser aging effects and lower accuracy. Therefore, we compromise between the visual quality of the synthesized and accuracy of aged images and decided to use $k=32$ for all our experiments. Note that, $k=0$, can achieve nearly the same accuracy as the feature aging module alone, however, the visual quality is compromised. Therefore, an appropriate $k$ can be chosen, depending on the security concerns and application.

\section{Implementation Details}
All models are implemented using Tensorflow r1.12.0. A single NVIDIA GeForce RTX 2080 Ti GPU is used for training and testing.

\subsection{Data Preprocessing} All face images are passed through MTCNN face detector~\cite{mtcnn} to detect five landmarks (two eyes, nose, and two mouth corners). Via similarity transformation, the face images are aligned. After transformation, the images are resized to $160\times 160$ and $112\times 96$ for FaceNet~\cite{facenet} and CosFace~\cite{cosface}, respectively.

\paragraph{Feature Aging Module} For all the experiments, we stack two fully connected layers and set the output of each layer to be of the same $d$ dimensionality as the ID encoder's feature vector.

\paragraph{Image Generator} 
All face images are cropped and aligned via MTCNN~\cite{mtcnn} and resized to $160\times160$. The style-encoder is composed of four convolutional layers and a fully connected layer in the last stage that outputs a $k$-dimensional style feature vector. In all our experiments, $k=32$. The decoder first concatenates the $k$-dimensional style vector and the $d$-dimensional ID vector from the ID encoder into a ($k+d$)-dimensional vector followed by four-strided convolutional layers that spatially upsample the features. All convolutional and strided convolutional layers are followed by instance normalization with a leaky ReLU activation function. At the end, the decoder outputs a $160\times160\times3$ image (for FaceNet) and $112\times96\times3$ image (for CosFace).
We emperically set $\lambda_{ID} = 1.0$, $\lambda_{pix}=10.0$, and $\lambda_{tv}=1e-4$.


We train the proposed framework for 200,000 iterations with a batch size of 64 and a learning rate of $0.0002$ using Adam optimizer with parameters $\beta_1 = 0.5, \beta_2 = 0.99$. In all our experiments, $k=32$.


\section{Visualizing Feature Aging}
\begin{figure*}[!h]
\centering
    \includegraphics[width=\linewidth]{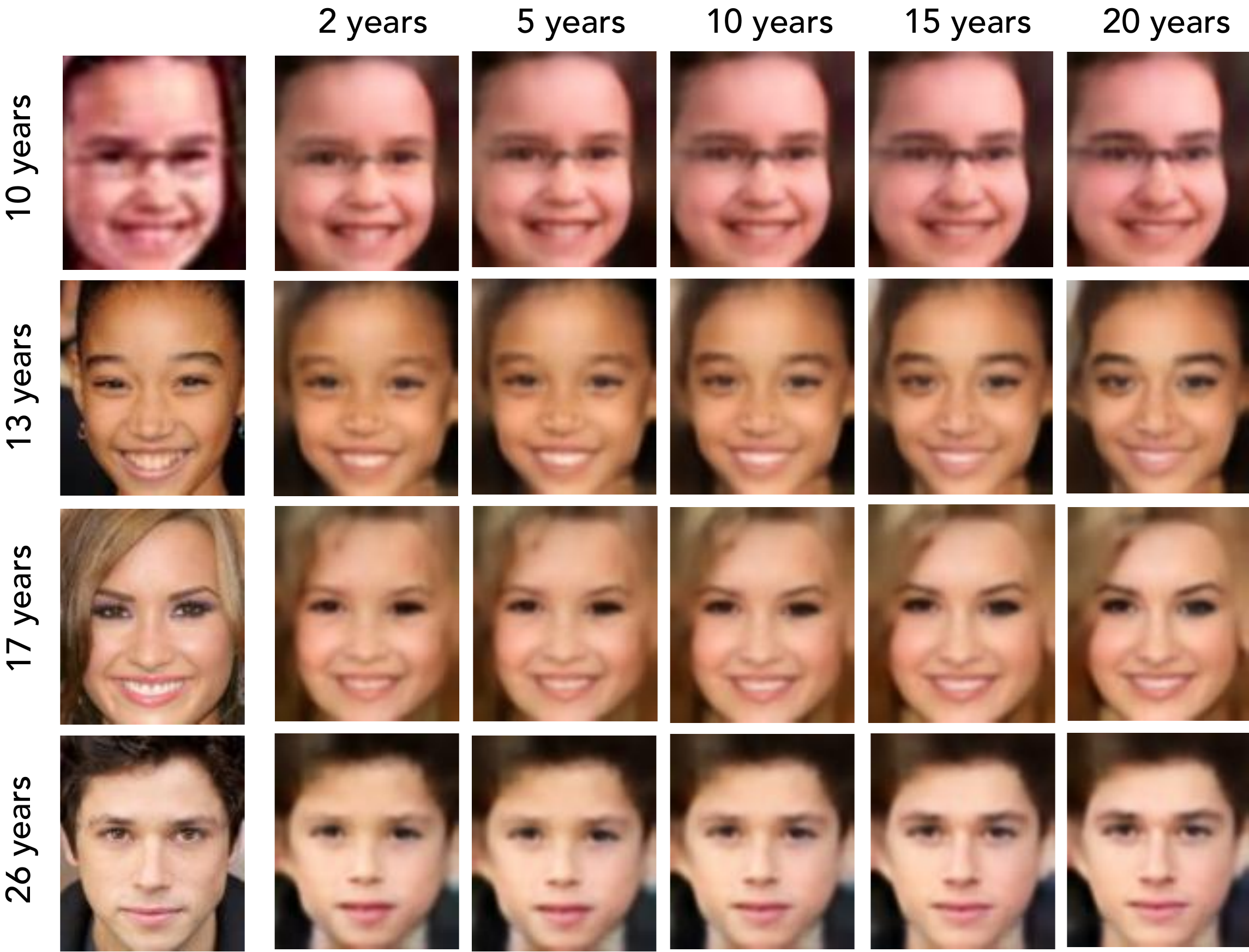}
    \caption{Aged face images from the proposed aging scheme using CosFace\cite{cosface} to specified target ages on ITWCC dataset.}
    \label{fig:examples_ITWCC}
\end{figure*}

\begin{figure*}[!h]
\centering
    \includegraphics[width=\linewidth]{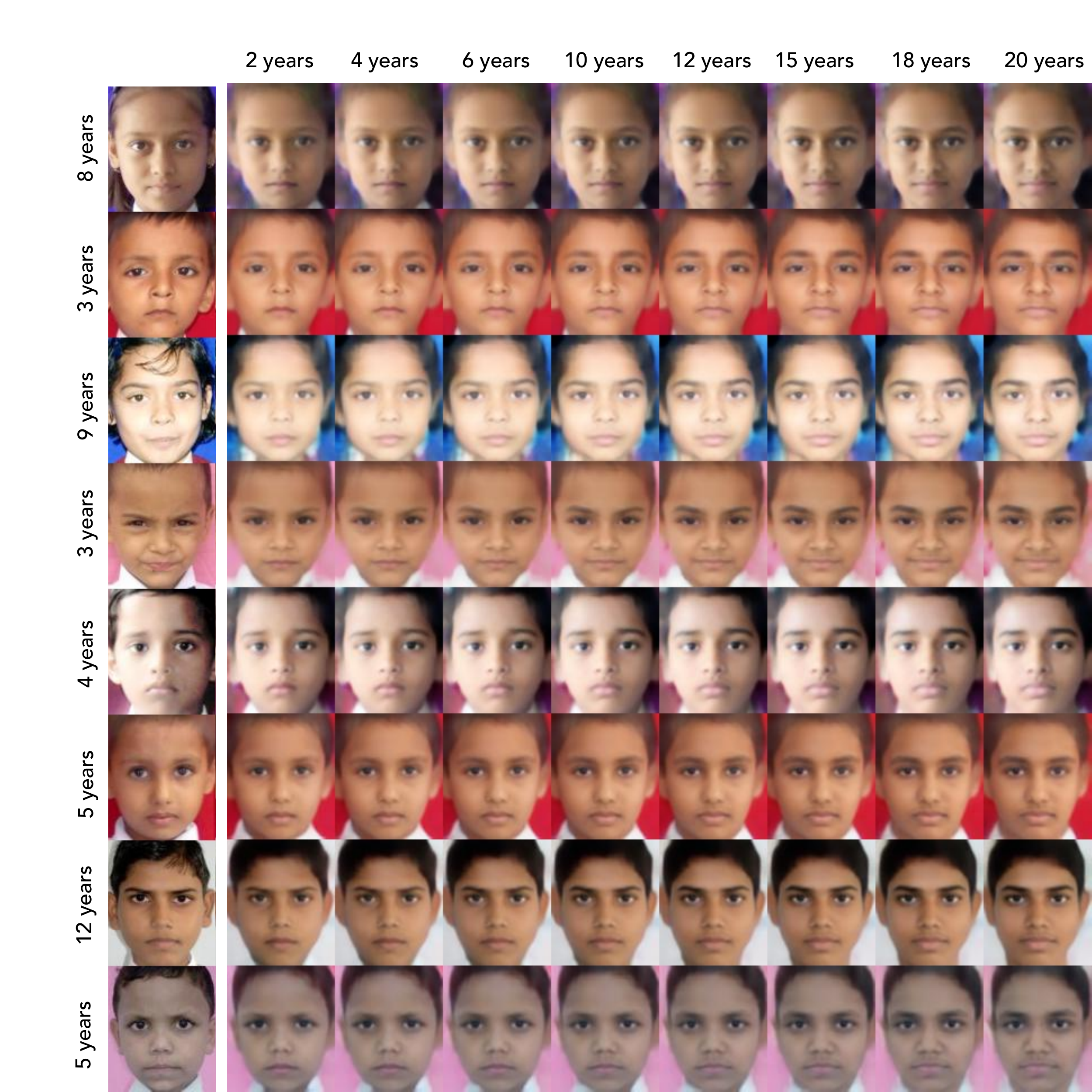}
    \caption{Aged face images from the proposed aging scheme using CosFace\cite{cosface} to specified target ages on CFA dataset.}
    \label{fig:examples_CLF}
\end{figure*}

In Figures~\ref{fig:examples_ITWCC} and~\ref{fig:examples_CLF}, we plot additional aged images via our proposed aging scheme to show the effectiveness of our feature aging module and image generator.

\end{document}